\documentclass[10pt]{article}

\usepackage[a4paper,margin=1.0in]{geometry}
\usepackage[T1]{fontenc}
\usepackage[utf8]{inputenc}
\usepackage{microtype}

\usepackage{amsmath,amssymb,amsthm}
\usepackage{mathtools}

\usepackage{stix2}

\usepackage{graphicx}
\usepackage{booktabs}
\usepackage[table]{xcolor}
\usepackage[colorlinks=true,
            linkcolor=IFMTealBlue,
            citecolor=IFMBlue,
            urlcolor=IFMBlue,
            filecolor=IFMTealBlue]{hyperref}
\usepackage{titlesec}
\usepackage{fancyhdr}
\usepackage{tikz}
\usepackage{ragged2e}
\usepackage{array}
\usepackage{caption}
\usepackage[hang,flushmargin]{footmisc}

\usepackage{natbib}

\usepackage{multirow}
\usepackage{float}
\usepackage{makecell}
\usepackage{algorithm}
\usepackage{algorithmic}
\usepackage{rotating}
\usepackage{adjustbox}
\usepackage{lineno}
\usepackage{xspace}
\usepackage{pifont}
\usepackage{bm}
\usepackage{nicefrac}
\usepackage{url}
\usepackage{cellspace}
\usepackage{csquotes}
\usepackage{empheq}
\usepackage{multicol}
\usepackage{wrapfig}
\usepackage{threeparttable}
\usepackage{pdflscape}
\usepackage[most]{tcolorbox}
\tcbuselibrary{breakable,skins}

\newtcbox{\mymath}[1][]{
    nobeforeafter, math upper, tcbox raise base,
    enhanced, colframe=blue!30!black,
    colback=blue!30, boxrule=1pt,
    #1}
\usepackage{enumitem}

\definecolor{TextBlack}{HTML}{111111}

\definecolor{IFMDarkBlue}{HTML}{284269}
\definecolor{IFMTealBlue}{HTML}{34a292}
\definecolor{IFMLightTealBlue}{HTML}{ecfaf8}
\definecolor{IFMBlue}{HTML}{52699F}

\definecolor{CoolSlate}{HTML}{6F7D88}
\definecolor{distillcolor}{RGB}{220,53,69}
\definecolor{sftcolor}{RGB}{40,167,69}
\definecolor{ourscolor}{RGB}{0,123,255}
\definecolor{annotcolor}{HTML}{6F7D88}         % CoolSlate
\definecolor{questionbg}{RGB}{255,248,225}
\definecolor{toolbg}{HTML}{E8EBED}             % CoolSlate!10
\definecolor{planbg}{RGB}{232,241,255}

\definecolor{hookersgreen}{rgb}{0.0, 0.44, 0.0}
\definecolor{indiagreen}{rgb}{0.07, 0.53, 0.03}
\definecolor{islamicgreen}{rgb}{0.0, 0.56, 0.0}
\definecolor{kellygreen}{rgb}{0.3, 0.73, 0.09}
\definecolor{alizarin}{rgb}{0.82, 0.1, 0.26}
\definecolor{goldenrod}{RGB}{218,165,32}

\definecolor{mygreen}{HTML}{4A9E5C}
\definecolor{myorange}{HTML}{E8913A}
\definecolor{myblue}{HTML}{2171B5}
\definecolor{mylightgreen}{HTML}{9DD4A5}
\definecolor{myred}{HTML}{C43E3E}

\definecolor{stepbg}{HTML}{EEF1F3}              % CoolSlate!5
\definecolor{planblue}{RGB}{0,123,255}
\definecolor{correctgreen}{RGB}{40,167,69}
\definecolor{stepbanner}{HTML}{6F7D88}          % CoolSlate
\definecolor{reasoncolor}{HTML}{6F7D88}         % CoolSlate
\definecolor{tagcolor}{RGB}{0,100,160}
\definecolor{plancolor}{RGB}{0,90,50}
\definecolor{answergreen}{RGB}{0,120,60}
\definecolor{questioncolor}{RGB}{0,0,0}

\definecolor{Gray}{gray}{0.95}

\color{TextBlack}

\makeatletter
\g@addto@macro\bfseries{\color{IFMDarkBlue}}
\makeatother

\let\oldtextsuperscript\textsuperscript
\renewcommand{\textsuperscript}[1]{{\color{IFMTealBlue}\oldtextsuperscript{#1}}}

\makeatletter
\renewcommand\@makefnmark{{\color{IFMDarkBlue}\hbox{\@textsuperscript{\normalfont\@thefnmark}}}}
\makeatother

\newcommand{\affmark}[1]{\textsuperscript{#1}}

\titleformat{\section}
  {\large\bfseries}
  {\thesection}{0.7em}{}
\titleformat{\subsection}
  {\normalsize\bfseries}
  {\thesubsection}{0.6em}{}
\titleformat{\subsubsection}
  {\normalsize\bfseries}
  {\thesubsubsection}{0.5em}{}

\titlespacing*{\section}{0pt}{1.25em}{0.45em}
\titlespacing*{\subsection}{0pt}{0.9em}{0.3em}
\titlespacing*{\subsubsection}{0pt}{0.7em}{0.25em}

\newenvironment{abstractpanel}
  {\begin{center}
   \begin{tikzpicture}
   \node[
     fill=IFMLightTealBlue,
     draw=none,
     rounded corners=7pt,
     inner xsep=12pt,
     inner ysep=10pt,
     text width=\dimexpr\linewidth-24pt\relax,
     align=justify
   ] \bgroup
  }
  {
   \egroup;
   \end{tikzpicture}
   \end{center}
  }

\DeclareMathOperator*{\argmax}{arg\,max}

\tcbset{
    questionbox/.style={
        breakable,
        colback=questionbg, colframe=CoolSlate!60, fonttitle=\bfseries,
        boxrule=0.5pt, arc=2pt, left=4pt, right=4pt, top=2pt, bottom=2pt,
        before skip=6pt, after skip=4pt
    },
    distillbox/.style={
        breakable,
        colback=distillcolor!4, colframe=distillcolor!50, fonttitle=\bfseries\color{distillcolor},
        boxrule=0.5pt, arc=2pt, left=4pt, right=4pt, top=2pt, bottom=2pt,
        before skip=4pt, after skip=2pt
    },
    sftbox/.style={
        breakable,
        colback=sftcolor!4, colframe=sftcolor!50, fonttitle=\bfseries\color{sftcolor},
        boxrule=0.5pt, arc=2pt, left=4pt, right=4pt, top=2pt, bottom=2pt,
        before skip=4pt, after skip=2pt
    },
    oursbox/.style={
        breakable,
        colback=ourscolor!4, colframe=ourscolor!50, fonttitle=\bfseries\color{ourscolor},
        boxrule=0.5pt, arc=2pt, left=4pt, right=4pt, top=2pt, bottom=2pt,
        before skip=4pt, after skip=2pt
    },
    annotbox/.style={
        breakable,
        colback=CoolSlate!8, colframe=CoolSlate!40, fonttitle=\bfseries,
        boxrule=0.4pt, arc=2pt, left=4pt, right=4pt, top=2pt, bottom=2pt,
        title={Observation}, before skip=4pt, after skip=8pt
    },
    fpquestion/.style={
        breakable,
        colback=questionbg, colframe=CoolSlate!60, fonttitle=\bfseries,
        boxrule=0.5pt, arc=2pt, left=4pt, right=4pt, top=2pt, bottom=2pt,
        before skip=6pt, after skip=4pt
    },
    fpstep/.style={
        breakable,
        colback=stepbg, colframe=CoolSlate!30,
        boxrule=0.3pt, arc=1pt, left=3pt, right=3pt, top=1pt, bottom=1pt,
        before skip=2pt, after skip=1pt
    },
    fpanswer/.style={
        breakable,
        colback=correctgreen!8, colframe=correctgreen!50, fonttitle=\bfseries,
        boxrule=0.5pt, arc=2pt, left=4pt, right=4pt, top=2pt, bottom=2pt,
        before skip=4pt, after skip=6pt
    },
    trajectory/.style={
        breakable, enhanced,
        colback=white, colframe=CoolSlate!70,
        coltext=reasoncolor,
        boxrule=0.6pt, arc=3pt,
        left=5pt, right=5pt, top=4pt, bottom=4pt,
        fonttitle=\bfseries\small,
        before skip=8pt, after skip=8pt,
    },
    stepbanner/.style={
        colback=stepbanner, colframe=stepbanner,
        boxrule=0pt, arc=0pt,
        left=5pt, top=2pt, bottom=2pt,
        before skip=6pt, after skip=4pt,
        fontupper=\bfseries\small\color{white}
    },
}

\usepackage{tikz}

\newcommand{\modelname}{{{\textsc{SiRA}}}\xspace}

\usepackage{color-edits}
\addauthor{gn}{magenta}

\DeclareUnicodeCharacter{202F}{\,}

\newcommand{\papertitle}{General Agentic Planning Through \\ Simulative Reasoning with World Models}

\newcommand{\paperauthors}{%
  Mingkai Deng\affmark{1,2,*},
  Jinyu Hou\affmark{1,2,*},
  Zhiting Hu\affmark{1,3},
  Eric P.~Xing\affmark{1,2}
}

\newcommand{\paperaffiliations}{%
  \affmark{1} Institute of Foundation Models (IFM) \quad
  \affmark{2} Carnegie Mellon University \quad
  \affmark{3} UC San Diego
}

\newcommand{\papernote}{%
  \affmark{*}Co-First Author \hspace{1em}|\hspace{1em}
  Contact:
  \href{mailto:mingkaid@cs.cmu.edu}{\{mingkaid,jinyuhou\}@cs.cmu.edu}
}

\newcommand{\makepreprinttitle}{%
  \begin{center}
    \vspace*{-2.9em}  

    {\fontsize{17}{22}\selectfont\bfseries \papertitle\par}

    \vspace{0.7em}
    {\normalsize\bfseries \paperauthors\par}

    \vspace{0.2em}
    {\small \paperaffiliations\par}

    \vspace{0.1em}
    {\small \papernote\par}
  \end{center}
}

\begin{document}

\makepreprinttitle

\begin{abstractpanel}
\noindent\textbf{Abstract.}
What does it mean to plan? How should an intelligent agent reason about its actions for decision-making across diverse tasks and environments?
Current agentic systems, whether built on scaffolded workflows or end-to-end trained policies, predominantly rely on reactive decision-making: selecting the next action using a fixed procedure (e.g., neural network, workflow), with at most undifferentiated adaptive computation (e.g., chain-of-thought) that lacks explicit modeling of future outcomes.
This reactive paradigm limits generalizability, as each new task or environment demands re-engineering rather than transfer of a shared reasoning capacity.
Humans, by contrast, plan by mentally simulating the consequences of candidate actions within an internal model of the world, a capacity known as \emph{simulative reasoning} (System~II) that supports flexible, goal-directed behavior across diverse contexts.
In this paper, we argue that simulative reasoning through a world model provides a general-purpose planning mechanism for agentic systems, improving upon reactive policies (System~I) by grounding decisions in predicted future states rather than pattern-matched responses.
To verify this hypothesis, we introduce \modelname (\underline{Si}mulative \underline{R}easoning \underline{A}rchitecture), a goal-oriented architecture that instantiates simulative reasoning using an LLM-based world model with natural-language belief states, while remaining model-agnostic in design.
We evaluate across three qualitatively distinct task categories: constrained navigation, multi-hop information aggregation, and general instruction following, instantiated in a web-browser environment.
Across all categories, simulative reasoning achieves up to 124\% higher task completion rates than a matched reactive baseline, and increases the success rate on constrained navigation from 0\% to 32.2\% compared to a representative open-web agent.
The persistent advantage across distinct task types suggests that the benefit stems from generalizable counterfactual evaluation rather than task-specific tuning.\footnote{\raggedright All experiments were conducted in 2024 with then-available models and tooling.}
We release the web-browsing agent built on \modelname as an open-source research artifact.\footnote{\url{https://github.com/sailing-lab/sira}}

\end{abstractpanel}

\section{Introduction}

\begin{figure}
    \centering
    \includegraphics[width=0.7\linewidth]{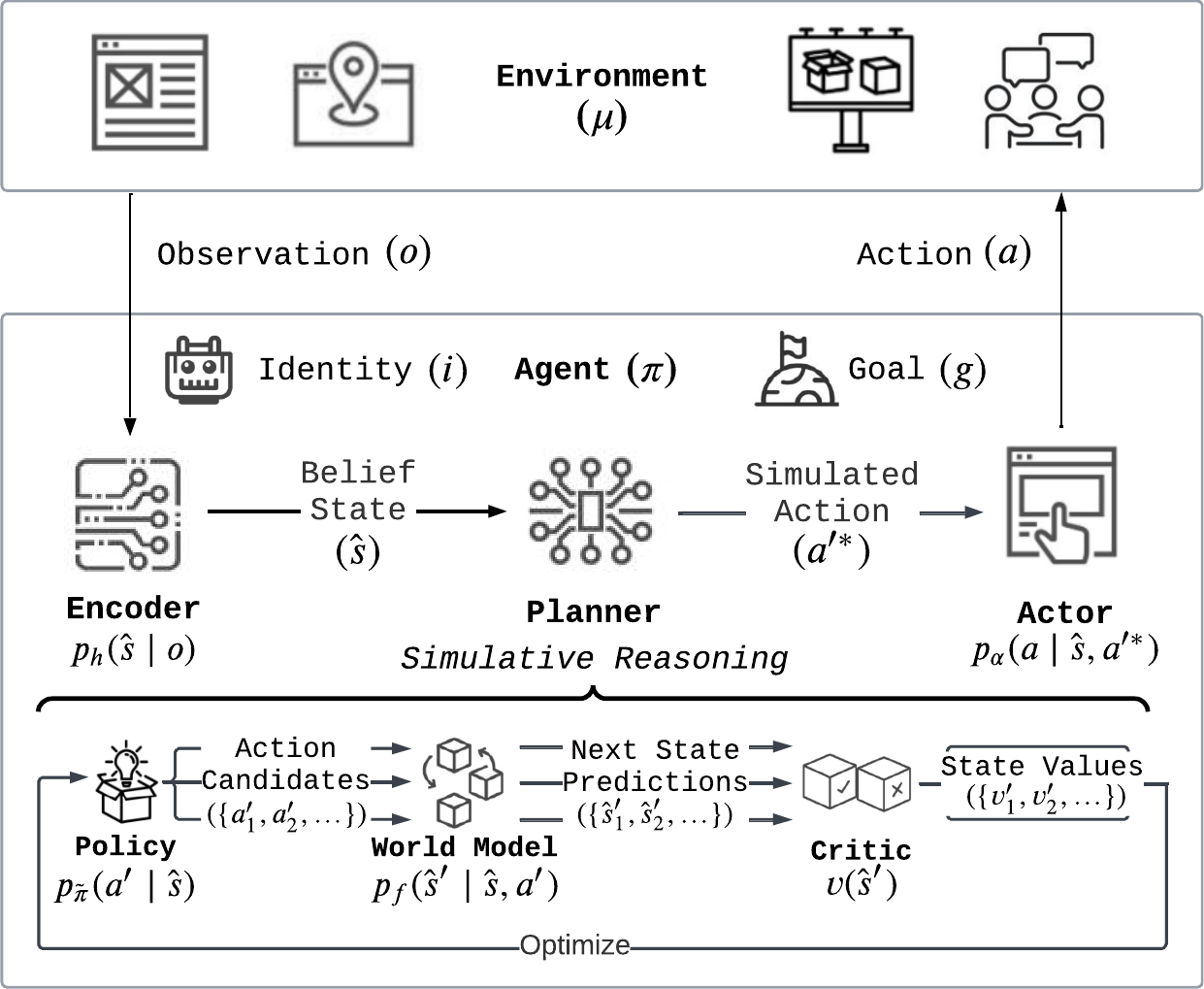}
    \caption{Instantiation of simulative reasoning in \modelname. At each time step, the encoder $h$ maps the observation $o$ into a natural-language belief state, based on which the planner proposes candidate actions, simulates their consequences through the world model $f$, evaluates goal progress via the critic $v$, and passes the best simulated action to the actor $\alpha$ for execution.}
    \label{fig:design-using-llm}
\end{figure}

Research in AI has long aimed to create agents that can plan and pursue goals effectively over long time horizons~\citep{mccarthy1955dartmouth,newell1959report}.
Contemporary agentic systems have achieved impressive results across web and computer automation~\citep{openai_cua_2025,zhou2023webarena}, software development~\citep{anthropic2025claudecode,anysphere2025cursor}, and scientific research~\citep{openai_learning_to_reason,guo2025deepseek,gottweis2025towards}, often by interacting with carefully designed harnesses of environments, tools, and predefined interaction logic~\citep[e.g.,][]{anthropic_mcp_2024}.

Despite their practical success, these systems share a common structural limitation: their decision-making is fundamentally \emph{reactive}.
Whether the system is a scaffolded LLM pipeline, a prompt-based workflow, or an end-to-end trained policy, it selects the next action using a fixed procedure (e.g., neural network, workflow) ---
% without explicitly modeling or evaluating future states.
much of their capability arises from external engineering (e.g., predefined tools, workflows, and interaction protocols) rather than from internal deliberation~\citep{xing2025critiques}.
When they do reason internally, planning is expected to emerge implicitly from undifferentiated chain-of-thought~\citep{wei2022chainofthought,yao2023react},%
\footnote{We use ``reactive policy'' (System~I) to denote the procedural mode of decision-making that iteratively predicts the next abstract internal state $z_t$ based on $p(z_t \mid z_{<t}, x)$ given input $x$, without explicit modeling or simulation of future outcomes. Chain-of-thought reasoning in LLMs is one instance; learned latent representations in end-to-end policies are another.}
a mode of adaptive computation that may extend the depth of processing but provides no mechanism for proposing, simulating, and comparing candidate futures.
Recent reasoning LLMs optimized via end-to-end RL~\citep{openai_learning_to_reason,guo2025deepseek} have scaled this adaptive computation substantially, yet their reasoning remains structurally reactive, with the main difference being generating longer chains of thought that better fit the training data.
As a result, even strong models remain vulnerable to locally myopic decisions and error accumulation over extended trajectories~\citep{andreas2022languagemodelsagentmodels,su2025underthinking}.
We argue that this results from the conflation of two distinct concepts: \textbf{internal compute} (generating more tokens or richer latent representations to better fit the observation) and \textbf{planning} (proposing candidate actions, predicting their consequences through a model of the environment, and selecting behavior grounded in these predictions).
A system can learn to produce useful intermediate reasoning without possessing the core primitive that planning requires: a grounded way to reason about counterfactual dynamics --- \emph{what would happen if we took action $a$ from state $s$?}

Humans, in contrast, adapt to diverse tasks and environments not only through fast, instinctive reactions (System~I), but by imagining potential outcomes, simulating possibilities using a mental model of the world, and planning accordingly (System~II)~\citep{kahneman2011thinking,Ball_2020}.
We refer to this deliberative capacity as \emph{simulative reasoning}: the ability to internally model future states resulting from candidate actions, evaluate their consequences, and select behavior grounded in these predictions.
This capacity has deep roots in sequential decision-making, from model-predictive control~\citep{camacho2007model} and learned world models~\citep{schrittwieser2020mastering,hafner2019planet} to recent work arguing its centrality for flexible, goal-directed behavior~\citep{ha2018world,lecun2022pathtowards,xing2025critiques}.
Crucially, simulative reasoning is a \emph{general-purpose planning mechanism}: because it operates by simulating state transitions rather than pattern-matching domain-specific responses, it can in principle transfer across tasks and environments without re-engineering.

In this paper, we argue that augmenting reactive policies (System~I) with simulative reasoning through a world model (System~II) provides a generalizable improvement to agentic planning, and we verify this hypothesis empirically.
We introduce \modelname (Simulative Reasoning Architecture), a goal-oriented architecture that instantiates this principle (Figure~\ref{fig:design-using-llm}).
\modelname introduces a \emph{world model} as the core engine for planning via simulation, constructing explicit simulative plans $c_t = (\hat{s}_t, a'_t, \hat{s}_{t+1}, a'_{t+1}, \dots, \hat{s}_{T'})$ consisting of proposed actions interleaved with predicted future states.
To make simulation tractable, \modelname represents belief states as compact natural-language summaries, leveraging discrete, concept-based representations for robust reasoning.
To enhance adaptability across environments, it adopts a hierarchical architecture that isolates perception, simulative planning, and action selection.
While we implement these components using LLMs as a substrate, the architecture is model-agnostic: any system capable of encoding observations, predicting state transitions, and evaluating outcomes can serve as the underlying world model.
Recent work has shown that LLMs can themselves serve as effective world models in the language space~\citep{hu2023language,hao2023reasoninglanguagemodelplanning}, offering a practical path toward this end.

To test whether simulative reasoning provides a generalizable advantage, we evaluate \modelname across three qualitatively distinct task categories: \emph{constrained navigation}, \emph{multi-hop information aggregation}, and \emph{general instruction following}, instantiated in a web-browser environment, which provides realistic complexity, partial observability, and long-horizon interaction.
To enable controlled evaluation under constrained navigation, we develop FlightQA (Section~\ref{sec:flightqa}), a dataset of flight search questions with systematically varying constraint complexity.
Results show that simulative reasoning (System~II) outperforms a matched reactive baseline (System~I) by up to \textbf{124\%} in task-completion rate, and increases the success rate on constrained navigation from \textbf{0\%} to \textbf{32.2\%} compared to a representative open-web agent.
The persistent advantage across qualitatively distinct task types provides initial evidence that the benefit stems from generalizable counterfactual evaluation rather than task-specific tuning.
All experiments were conducted in 2024 with the then-available models and browser tooling.
The code is available at \url{https://github.com/sailing-lab/sira}

\section{Related Work}

\paragraph{Reactive Agentic Systems}
The dominant paradigm in current agent design relies on reactive policies, where the agent selects the next action based on the current observation without explicit simulation of future outcomes.
This paradigm spans a broad range of implementations.
One approach focuses on data collection in the targeted environment followed by model training, as in AutoWebGLM~\cite{lai2024autowebglmlargelanguagemodelbased}, AgentQ~\cite{putta2024agentqadvancedreasoning}, and UI-TARS~\cite{qin2025uitarspioneeringautomatedgui}.
Prompt-based workflows offer an alternative, with systems like AWM~\cite{wang2024agentworkflowmemory} and VOYAGER~\cite{wang2023voyageropenendedembodiedagent} achieving strong results through carefully designed modules.
Proprietary agents such as OpenAI's Operator~\cite{openai_cua_2025}, Anthropic's Computer Use~\cite{anthropic2024claude35sonnet}, and Google-DeepMind's Project Mariner~\cite{deepmind2024mariner}, along with open-source systems including OpenHands BrowsingAgent~\cite{openhands}, WebVoyager~\cite{he2024webvoyagerbuildingendtoendweb}, CogAgent~\cite{hong2024cogagentvisuallanguagemodel}, and WebAgent~\cite{gur2024realworldwebagentplanninglong}, are typically built on ReAct-based~\cite{yao2023react} reasoning.
Multi-agent workflows~\cite{hu2025owloptimizedworkforcelearning, jiang2025longtermmemoryfoundation, zhang2025agentorchestrahierarchicalmultiagentframework, fourney2024magenticonegeneralistmultiagentsolving} and code-action systems~\cite{wang2024executablecodeactionselicit, openhands, smolagents, qiu2025alitageneralistagentenabling} extend the scope of reactive agents by delegating to specialists or composing executable scripts.
Despite their diversity in domain and implementation, these systems share the structural limitation that behavior is generated without explicitly modeling or evaluating future states: the agent reacts to what it observes rather than reasoning about what would happen next.

\paragraph{World Models and Model-Based Planning}
Model-based planning, in which an agent uses a predictive model to simulate the consequences of candidate actions before committing to one, has a long history in sequential decision-making.
Early successes demonstrated the approach in classic games such as Go, chess, shogi, and Atari~\cite{oh2015actionconditionalvideopredictionusing, Schrittwieser_2020}, where the environment dynamics are known or learnable.
Predictive models were subsequently applied to policy optimization in continuous control tasks~\cite{janner2021trustmodelmodelbasedpolicy, hansen2022temporaldifferencelearningmodel}.
More recently, with the increasing capabilities of foundation models, model-based planning has been extended to more complex problems including mathematical reasoning~\cite{hao2023reasoninglanguagemodelplanning}, open-ended game playing~\cite{hafner2024masteringdiversedomainsworld}, and web browsing~\cite{gu2025llmsecretlyworldmodel}.
However, prior work in this space typically represents world states using holistic continuous embeddings, which can suffer from noise and high variability that detracts from robust decision-making~\cite{barrett2017emotions}, and targets narrow domains rather than treating simulative reasoning as a general-purpose planning pattern.

\paragraph{Benchmarks and Evaluation}
Numerous benchmarks have been introduced to evaluate agentic systems, including WebArena~\cite{zhou2023webarena},
WebVoyager~\cite{he2024webvoyagerbuildingendtoendweb},
MiniWoB++~\cite{liu2018reinforcement},
Mind2Web~\cite{deng2023mind2webgeneralistagentweb}, and WebShop~\cite{yao2023webshopscalablerealworldweb}.
Despite wide adoption, these benchmarks are usually either built in simulated and simplified environments, based on outdated questions, or lack convincing methods of measuring task completion, which detracts from the goal of evaluating practically useful agents.
To address these challenges, we build FlightQA, a dataset for evaluating agent reasoning under systematically varying constraint complexity in live web environments. More details are included in Section~\ref{sec:flightqa}.

\section{Simulative Reasoning for Agentic Planning}
\subsection{Formulation of Agent-Environment Model}
\begin{figure}
    \centering
    \includegraphics[width=0.85\linewidth]{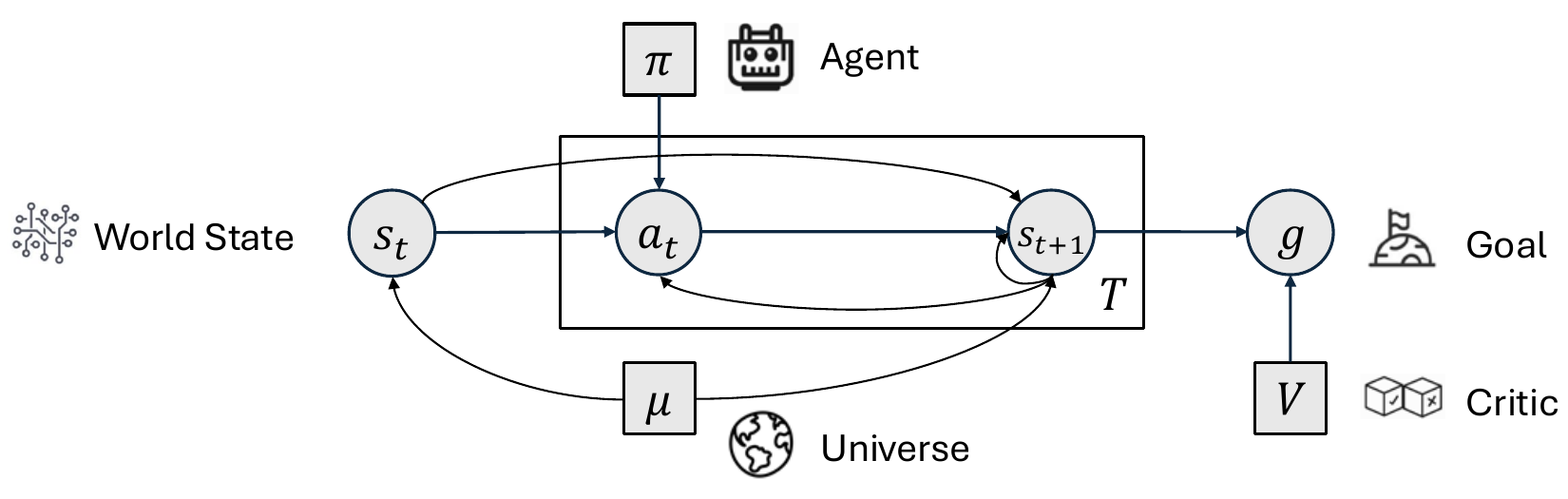}
    \caption{A possible definition of an optimal agent}
    \label{fig:optimal-agent}
\end{figure}

We first present the formulation of an optimal goal-oriented agent following the agent-environment model presented in~\cite{xing2025critiques}:
We consider an agent $\pi$ with identity $i$ (e.g., name, description) and goal $g$ acting in environment $\mu$ (e.g., web browser, physical world, the entire universe) with action space $\mathcal{A}$ and state space $\mathcal{S}$ (Figure~\ref{fig:optimal-agent}). Formally, at each time step $t$, the \textbf{agent} $\pi$ takes the current state $s_t \in \mathcal{S}$ and outputs the next action $a_t \in \mathcal{A}$ following a policy distribution $p_{\pi} (a_t \mid s_t)$, while the \textbf{environment} $\mu$ takes the current state $s_t$ and action $a_t$, and outputs the next state $s_{t+1} \in \mathcal{S}$ based on the distribution $p_{\mu}(s_{t+1} | s_t, a_t)$. 
We can thus denote the distribution of the interaction trajectory up to timestep $T$, or $(a_t, s_{t+1}, \dots, a_{T-1}, s_T)$ given the current state $s_t$, as below:
\begin{equation}
p^{\pi}_{\mu}(a_t, s_{t+1}, \dots, a_{T-1}, s_T \mid s_t) = \prod_{k=t}^{T-1} {\underbrace {\textstyle p_\pi(a_k \mid s_k)}_{\text{ agent }} } \ {\underbrace {\textstyle p_\mu(s_{k+1} \mid s_k, a_k)}_{\text{ environment }} }
\label{eq:trajectory}
\end{equation}
In each state $s_t$, the agent also receives a reward $r(g, s_t)$ based on its goal $g$. We evaluate the agent by its discounted cumulative reward, denoted as $\sum_{k=t}^\infty \gamma_k r(g, s_k)$ (with the discount parameter $\gamma_t$ decaying to zero with time, i.e., $\lim_{t \to \infty} \gamma_t = 0$). Note that this reward function can be dense (e.g., gaming scores), but perhaps frequently sparse (e.g., curing a disease). 
The agent's long-term success can thus be measured by its expected future discounted reward, also known as \textbf{value function}~\cite{sutton1998reinforcement}, which satisfies the following recurrence:
\begin{align}
    V_{\pi,\mu}^g(s_t) &\vcentcolon= \mathbb{E}_{\pi,\mu} \left[ \sum_{k=t}^\infty \gamma_k r(g, s_k) \mathrel{\bigg|} s_t \right] \nonumber \\
    &= \lim_{T \rightarrow \infty} \sum_{(a_t, s_{t+1},\dots, s_T)} \sum_{k=t}^T \gamma_k r(g, s_k) \ p^{\pi}_{\mu}(a_t, s_{t+1}, \dots, s_T \mid s_t) \nonumber \\
    &= \sum_{(a_t, s_{t+1},\dots, s_T)} \Bigg({\underbrace { \sum_{k=t}^{T-1} \gamma_k r(g, s_k) + \gamma_T V_{\pi, \mu}^g(s_T) }_{\text{goal progress}} }\Bigg) \ {\underbrace {\textstyle p^{\pi}_{\mu}(a_t, s_{t+1}, \dots, s_T \mid s_t)}_{ \text{trajectory} } },
    \label{eq:value-function}
\end{align}
Which indicates that the value function in state $s_t$ can be expressed in terms of the value function at possible future states $s_T$ weighted by their probabilities.

\subsection{Definition of Optimal Agent}
\label{subsec:optimal-agent}
 
Based on Equations~\ref{eq:trajectory} and \ref{eq:value-function}, we can define the optimal agent $\pi^*_\mu$ in environment $\mu$ as one that maximizes the value function, written formally as below: 
\begin{equation}
    \pi^*_\mu := \argmax_\pi V^{g}_{\pi,\mu}.
\end{equation} 
Some simple derivation will show that the optimal agent in state $s_t$ will follow the following decision rule $\pi^*_\mu$ when planning for actions $a_{t:T-1}$:
\begin{equation}
    \pi^*_\mu(s_t) = {\underbrace {\argmax_{ a_{t:T-1}} }_{\text{possible actions} } } \ \sum_{s_{t+1:T}}   \Bigg({\underbrace { \sum_{k=t}^{T-1} \gamma_k r(g, s_k) + \gamma_T V_{\pi, \mu}^g(s_T) }_{\text{goal progress}} }\Bigg) \prod_{i=t}^{T-1} \ {\underbrace {p_\mu(s_{i+1} \mid s_i, a_i)}_{\text{universe response}} } \
    \label{eq:optimal-decision-making-mu} 
\end{equation}
In practice, agents often samples promising action candidates using a policy function $\tilde{\pi}$ through the distribution $p_{\tilde{\pi}}(a_t \mid s_t)$. Building the optimal agent thus requires capabilities for proposing possible actions ($\tilde{\pi}$), predicting their outcomes ($\mu$), and evaluating goal progress ($r,V$), respectively. Note that typical \textbf{reactive agents} (System~I) that output the next action directly can be seen as taking the first sample from $\tilde{\pi}$, making fast, instinctive reactions~\cite{kahneman2011thinking}, without simulating and evaluating the outcomes using $\mu$ and $V$. \textbf{Simulative reasoning} (System~II), by contrast, performs deliberate decision-making by proposing multiple candidate actions, predicting their consequences, and selecting the best one, which provides opportunities for spotting and correcting errors through the deliberation process.

\subsection{World Model for Simulative Reasoning}
 
Indeed, simulative reasoning, which consists of proposing candidate actions, predicting their consequences through a model of the environment, and selecting behavior grounded in these predictions, provides a general-purpose planning mechanism applicable across environments and tasks~\cite{xing2025critiques}.
However, the optimal decision-making process defined in Equation~\ref{eq:optimal-decision-making-mu} requires the agent to have access to the ground-truth world state $s$ and the environment $\mu$ to experience and optimize over, which is often not available aside from simple scenarios like Go and Chess games~\citep{silver2016mastering,silver2017mastering}.
A \textbf{World Model} (WM) $f$ thus arises as a learned surrogate of the environment, enabling the agent to simulate consequences without direct interaction. 
Specifically, a WM $f$ operates on an internal representation of the world state, denoted as a \textit{belief state} $\hat{s}_t$, which is derived from sensory inputs $o_t$ via an \textit{Encoder} $h$ (unlike the optimal agent described in \S\ref{subsec:optimal-agent} which has direct access to the true world state $s_t$). 
Given proposed action $a_t$, the WM predicts the next belief state $\hat{s}_{t+1}$ according to the distribution $p_f(\hat{s}_{t+1} \mid \hat{s}_t, a_t)$. The predicted belief state then allows the agent to propose the next action, continuing the cycle of prediction and action up to the desired time horizon $T$. 
Thus, a WM essentially functions as a generative model of possible future world states, which enables simulative reasoning, or ``thought experiments". 
Formally, for the optimal agent $\pi^*_f$ equipped with WM $f$ in belief state $\hat{s}_t$, we define the simulation-based decision rule in Equation 6 as follows: 
\begin{equation}
    \pi^*_f(\hat{s}_t)= {\underbrace {\argmax_{ a_{t:T-1} } }_{\text{possible actions} } } \ \sum_{\hat{s}_{t+1:T}} \Bigg({\underbrace { \sum_{k=t}^{T-1} \gamma_k r(g, \hat{s}_k) + \gamma_{T}V_{\pi, f}^g(\hat{s}_{T}) }_{\text{goal progress}} }\Bigg)  \prod_{i=t}^{T-1} \ {\underbrace {p_f(\hat{s}_{i+1} | \hat{s}_i, a_i)}_{{\scriptsize \shortstack{simulation with\\world model}}}}
    \label{eq:world-model-decision-making}
\end{equation}
A general-purpose WM enables simulation of diverse possibilities across a wide range of domains, enabling agents to reason about outcomes without direct interaction with the environment. Because simulative reasoning operates by predicting state transitions rather than pattern-matching domain-specific responses, it can in principle transfer across tasks and environments, resulting in an advantage that reactive policies (System~I), however powerful, cannot replicate through increased internal compute alone.

\subsection{Instantiating Simulative Reasoning}
\label{subsec:agent-design}
The preceding subsection establishes simulative reasoning as a general planning principle. In this subsection, we describe two design decisions that enable robust and broad applicability across environments and tasks, including concept-based mixed representations and hierarchical planning, and then show how \modelname instantiates them.

\paragraph{Concept-Based Mixed Representations} 
Simulative reasoning requires the world model to predict how individual aspects of the state change in response to an action (e.g., ``the price field updated'' vs.\ ``the page layout shifted''). A factored, compositional state representation makes such targeted prediction tractable, but this is not available in a monolithic embedding.
Instead, the dominant approach to encoding observation $o_t$ (e.g., webpages, video streams) has been to directly map all input tokens into continuous embeddings with fixed dimensionalities $\hat{s}^z_t$.
While technically preserving all information, real-world sensory readings often suffer from inherent noise and high variability (e.g., ads on a webpage, varying weather and lighting conditions in video), which can make them brittle to reason over. Human cognition, on the other hand, has evolved to counter this variability by categorizing raw perception into \textit{discrete concepts}~\cite{barrett2017emotions}, which are often encoded in language, symbols or structured thoughts. Indeed, natural language is inherently hierarchical, capable of encoding concepts from concrete ones (e.g., apple) to highly abstract ones (e.g., religion). Discrete representations are also complete in general~\cite{xing2025critiques}, which ensures no information is necessarily lost in the compression process.
In \modelname, we instantiate this principle by representing the world state $\hat{s}_t$ using a discrete natural language summary $\hat{s}^c_t$ generated by a pretrained encoder model $h$, formally expressed as below:
\begin{equation}
    p_h(\hat{s}_t \mid o_t) = \prod_{i=1}^{N_t} p_h(\hat{s}_{t,i} \mid \hat{s}_{t,<i}, o_t),
\end{equation}
Where each $\hat{s}_{t,i}$ is a natural language token. Likewise, we also denote the WM $f$ which predicts the next state $\hat{s}_{t+1}$ as a natural language sequence $\hat{s}^c_{t+1}$, formally as below:
\begin{equation}
    p_f(\hat{s}_{t+1} \mid \hat{s}_t, a_t) = \prod_{i=1}^{N_{t+1}} p_h(\hat{s}_{t+1, i} \mid \hat{s}_{t+1,<i}, \hat{s}_t, a_t)
\end{equation}
This concept-based factorization allows the other modules (e.g., policy, critic) to operate on a structured latent space in which individual state components can be predicted and evaluated independently. In practice, we find this reduces hallucination and enables more robust planning, leading to better task performance.

\begin{figure}
    \centering
    \includegraphics[width=0.85\linewidth,page=2]{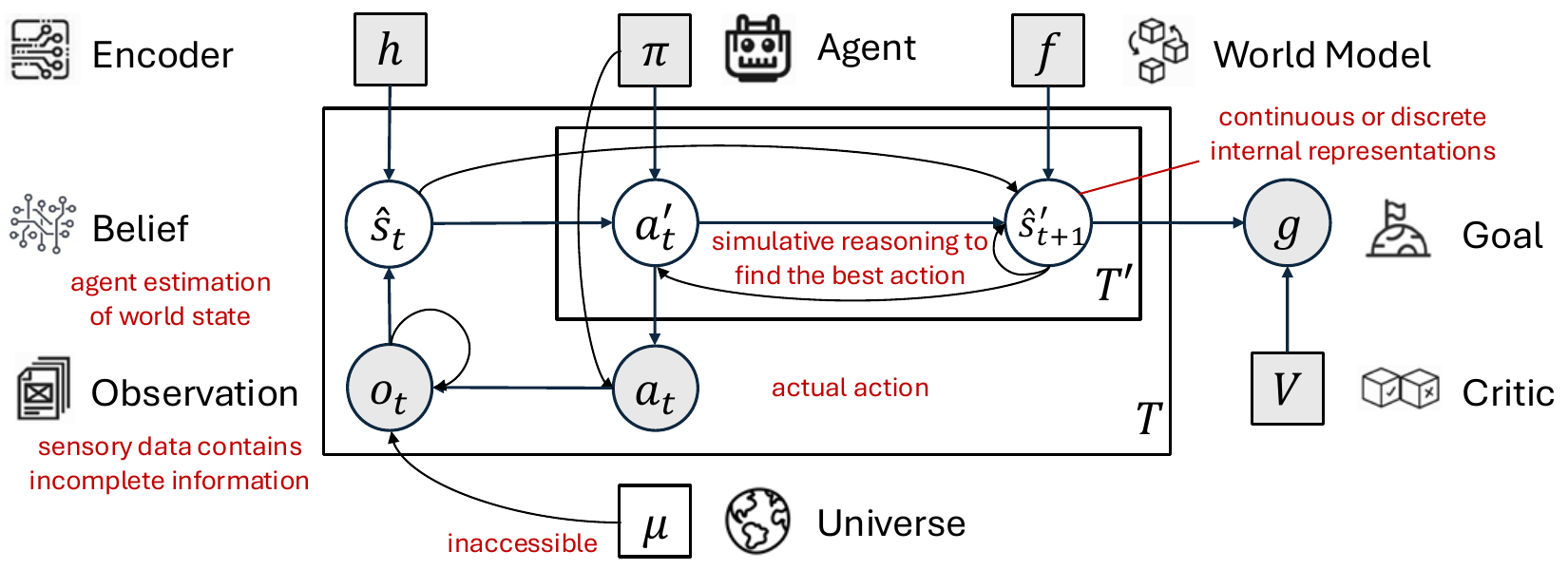}
    \caption{An agent in real world where groundtruth world state and universe are unavailable to experience or experiment, so world model is crucial for simulation. As discussed in \S\ref{subsec:agent-design}, separation of simulated actions $a_t'$ for planning and concrete actions $a_t$ for execution facilitates transfer and hierarchical planning, leading to more diverse and grounded actions which lead to better task success.}
    \label{fig:intro-world-model}
\end{figure}

\paragraph{Hierarchical Planning via Simulated Actions}
The customary approach to decision-making with world models has been to perform simulations or rollouts based on the specific action space $\mathcal{A}(\pi)$ afforded to the agent. While this approach indeed captures all the execution details, the specific idiosyncrasies of individual action spaces (e.g., parameter ordering, format, and scale) may hinder the transfer of knowledge across different action spaces, environments, and tasks, thereby limiting generalizable reasoning. Indeed, the real world may contain a richer range of intentions than what a particular action space offers (e.g., clicking on a flight may mean either exploring the pricing or committing to the option). Last but not least, the sequential roll-out over atomic actions can be inefficient and increase opportunities for error accumulation across multi-step, low-level predictions (e.g., swooshing of liquids with each muscle twitch), when higher-level dynamics over more abstract actions (e.g., spilling water due to tilting the glass) remain stable and predictable. 
To close this gap, we separate high-level, flexible planning from low-level, rigorous execution~\cite{sutton1991dyna}: the world model simulates over abstract actions that capture intentions, while a separate actor translates the chosen intention into concrete, environment-specific commands. In \modelname, this is realized as follows (Figure~\ref{fig:intro-world-model}): the agent's policy $p_{\tilde{\pi}}(a'_t \mid \hat{s}_t)$ and world model $p_f(\hat{s}_{t+1} \mid \hat{s}_t, a'_t)$ operate over simulated actions $a'_t$ from a separate action space $\mathcal{A}'$, while another actor $p_\alpha(a_t \mid a'_t, \hat{s}_t)$ is responsible for selecting the concrete action $a_t \in \mathcal{A}$ conditioned on the selected simulated action $a_t'$. 
This divide-and-conquer approach allows for more generalized reasoning disentangled from the exact details of the concrete action space and enables representation of a richer set of intentions. Furthermore, each simulated action $a_t'$ may represent multiple execution steps in the environment (e.g., ``explore the website'' vs ``click on the link''), which shortens the number of rollout steps for higher efficiency and fewer chances for error accumulation. 
In practice, we represent simulated actions $a_t'$ using natural language due to its generality and expressivity, and find it results in more diverse and grounded action proposals, leading to better task success.

\paragraph{Architecture and Decision Process}
Combining these design choices, we now describe the full decision process of \modelname: As illustrated in Figure~\ref{fig:design-using-llm}, given observation $o_t$ (e.g., webpage screenshots and/or accessibility tree), \modelname first infers the world state $\hat{s}_t$ using the encoder $h$, and then selects the best simulated action $a_t'^*$ through the planner. Inside the planner, the architecture performs simulative reasoning by proposing actions $a_t'$ using policy $\tilde{\pi}$ and predicting the next state $\hat{s}_{t+1}$ using the world model $f$, and evaluating goal progress $\sum_{k=t}^{T'-1} \gamma_k r(g, \hat{s}_k) + \gamma_{T'}V_{\pi, f}^g(\hat{s}_{T'})$ using critic $v$ upon reaching state $\hat{s}_{T'}$ at the planning horizon $T'$. This can repeat multiple times until the planner selects the action sequence $a'^*_{t:T'-1}$ with the highest expected success and passes the first step $a_t^*$ to actor $\alpha$ which finally outputs the concrete action $a_t$. 
Formally, \modelname can be seen as solving the following multi-level optimization problem:
\begin{align}
\hat{s}_{t} &= \argmax_{\hat{s}} \ {\underbrace {p_h(\hat{s} \mid o_t)}_{  \text{encoder} } } & \text{(Perception)} \nonumber  \\
a_{t:T'-1}'^* &= {\underbrace {\argmax_{ a_{t:T-1}' } }_{\scriptsize \shortstack{sampled from \\policy $\tilde{\pi}$}} } \sum_{\hat{s}_{t+1:T'}} \ {\underbrace { v(\hat{s}_{T'}) }_{\text{critic}} } \prod_{k=t}^{T'-1} {\underbrace {p_f(\hat{s}_{k+1} \mid \hat{s}_k, a_k')}_{{\scriptsize \shortstack{world model}}}} & \text{(Planning)} \\
a_t &= \argmax_{a} \ {\underbrace {p_\alpha(a \mid \hat{s}_t, a_t'^*)}_{ \begin{array}{c} \scriptsize \text{actor}\text{} \end{array} } }  & \text{(Acting)} \nonumber
\end{align}
For further clarity, we included the pseudo-code for \modelname's inference algorithm in Appendix~\ref{appendix:sira-inference}.

In \S\ref{sec:experiments} below, we present an instance of \modelname with each of these components implemented using pretrained LLMs. While these LLMs alone are often insufficient for many complex agentic tasks, \modelname's divide-and-conquer approach combines existing LLM strengths like instruction-following, summarization, reflection, and tool use to allow agentic behavior to emerge. Benefiting from massive web-scale pretraining on next-token prediction $p(x_t \mid x_{<t})$, which is formally akin to world modeling, LLMs possess significant potential to serve as world models with natural-language state and simulated action spaces~\cite{hao2023reasoninglanguagemodelplanning,hu2023language}.
We approximately infer the world state $\hat{s}_t$ and action $a_t$ by sampling from the LLM-based encoder and actor distributions $p_h$ and $p_\alpha$, respectively. For planning, we optimize over the sampled actions $a_{t:T'-1}'$ using readily available tree search algorithms like Depth-First Search (DFS) and Monte-Carlo Tree Search (MCTS).

\section{Experiments}
\label{sec:experiments}

We evaluate whether simulative reasoning (System~II) provides a generalizable advantage over reactive policies (System~I) across qualitatively distinct task categories. We select three categories that stress different aspects of agentic reasoning: \emph{constrained navigation}, \emph{multi-hop information aggregation}, and \emph{general instruction following}, and instantiate them in a web-browser environment, which provides realistic complexity, partial observability, long-horizon interaction, and multimodality~\citep{zhou2023webarena,gu2024your}. We choose the web browser as our testbed due to both its practical value and its technical challenge: it is an indispensable portal for individuals to perform many tasks in real life (e.g., gather information, book travels, submit applications), and it demands the agent to interact with diverse, dynamic interfaces under partial observability.
For each task category, we compare the following:
\begin{enumerate}
    \item \textbf{Simulative reasoning (System~II)}: The full \modelname architecture with world-model-based planning as described in \S\ref{subsec:agent-design}.
    \item \textbf{Reactive policy (System~I)}: \modelname with the planning module replaced by direct action selection from the policy, committing to the first sample without simulation. Formally: $a_t'^* = \argmax_{a_t'} p_{\tilde{\pi}}(a_t' \mid \hat{s}_t)$.
    \item \textbf{Unstructured Chain-of-Thought}: \modelname with the planning module replaced by strong reasoning LLMs that generate unstructured chain-of-thought (e.g., o1 and o3-mini). Due to the cost of running these models, we only report results on constrained navigation for reference.
    \item \textbf{BrowsingAgent} from OpenHands~\citep{openhands}, a representative open-web reactive agent which generates chain-of-thought before selecting an action.
\end{enumerate}
Comparing the reactive policy and simulative reasoning baselines within the same architecture isolates the effect of world-model-based simulation from other design choices (e.g., state representation, hierarchical action abstraction), providing a controlled test of the simulative reasoning hypothesis.

\begin{figure}
    \centering
    \includegraphics[width=0.8\linewidth]{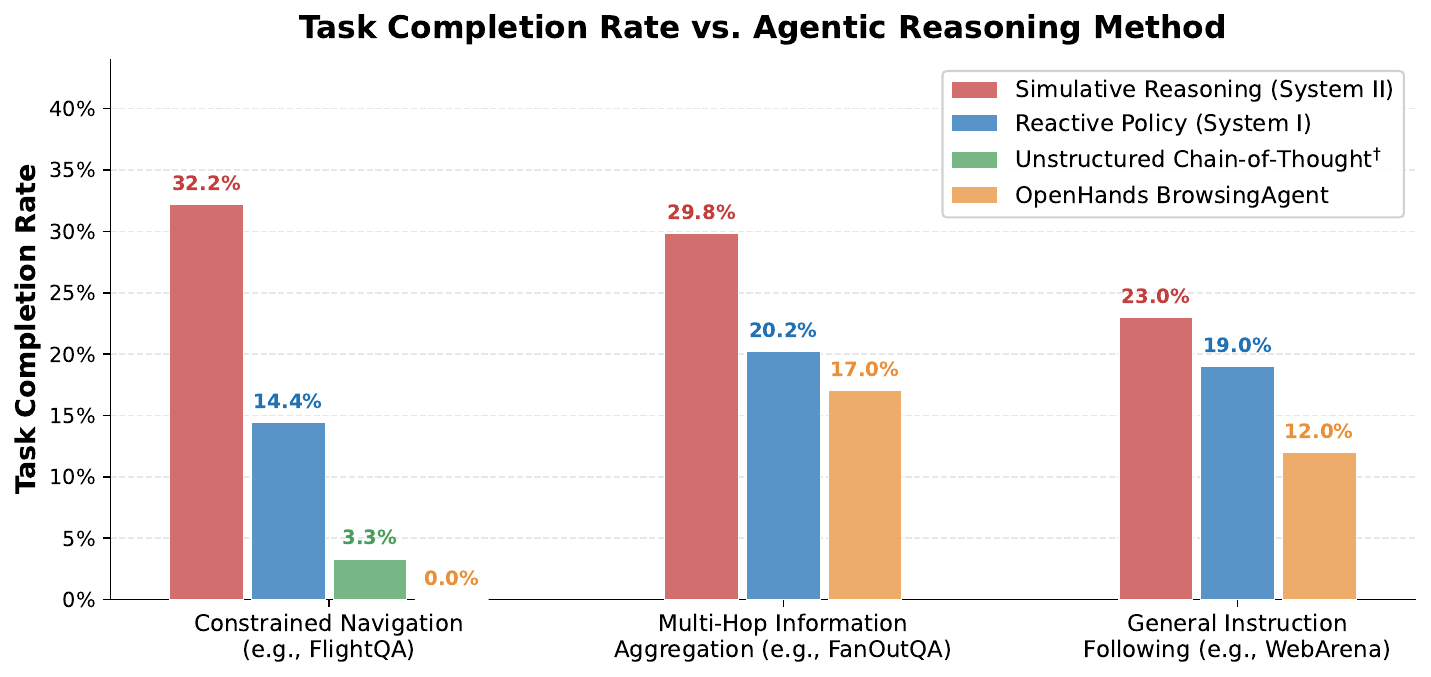}
    \caption{Task completion rate by reasoning method across three qualitatively distinct task categories. Simulative reasoning (System~II) achieves up to 124\% higher completion than the matched reactive policy (System~I), while unconstrained chain-of-thought, even using strong reasoning LLMs (e.g., o1, o3-mini), achieves near-zero success on constrained navigation.}
    \label{fig:results-overview}
\end{figure}

\paragraph{Implementation for Web Browsing}

We use prompts tailored to the web environments in this experiments, but plan to extend to other environments and move towards training a single agent model that can act optimally in a wider range of environments and tasks, which is an exciting next step. At each step $t$, the agent receives the observation $o_t$ as the HTML-based accessibility tree visible through the browser's viewport (an example is provided in Appendix~\ref{appendix:browsing-environment}). The agent then uses encoder LLM $h$ to summarize the observation as $\tilde{s}_t \sim p_h(\cdot \mid o_t)$, and then add it to a selective memory of past summaries and simulated actions $\{m(\tilde{s}_k, a_k'^*)\}_{k=1}^{t-1}$ to form the estimated world state $\hat{s}_t = [m(\tilde{s}_1, a_1'^*), \dots, m(\tilde{s}_{t-1}, a_{t-1}'^*), \tilde{s}_t]$ for planning. During planning, we sample $M$ simulated actions $a'_t$ from the policy $\tilde{\pi}$, cluster them into distinct actions, and use the world model $f$ to predict the next summary as $\tilde{s}_{t+1} \sim p_f(\cdot \mid \hat{s}_t, a_t')$ to form the next state $\hat{s}_{t+1} = [m(\tilde{s}_1, a_1'^*), \dots, m(\tilde{s}_t, a_t'), \tilde{s}_{t+1}]$; this repeats until the planning horizon $T$. To evaluate the terminal state $\hat{s}_T$ with critic $v$, we prompt the LLM to generate categorical answers and convert them into numerical scores (e.g., ``success'' receives a score of 1), and repeat for $N$ times to capture the fine-grained differences between states. Following previous work~\cite{koh2024tree,gu2025llmsecretlyworldmodel}, we set $M=N=20$ and $T=t+1$, and use DFS as the search algorithm. We implement the planning process using LLM Reasoners~\cite{hao2024llm}, a library for LLM-based complex reasoning using advanced algorithms. After the planner selects the simulated action $a_t'^*$, we update the memory with $m(\tilde{s}_t, a_t'^*)$.
For the actor $\alpha$, we additionally include the observation text $o_t$ in the prompt to ensure the action grounding. All the prompts are included in Appendix~\ref{appendix:prompts-browsing-implementation}.

\paragraph{Overview of Results}
An overview of our results is presented in Figure \ref{fig:results-overview}. Across all three task categories, simulative reasoning (System~II) uniformly outperforms both the reactive policy baseline (System~I) and the external BrowsingAgent baseline, increasing the success rate on constrained navigation from BrowsingAgent's 0\% to 32.2\%. Simulative reasoning also consistently improves over the matched reactive baseline by up to 124\% in task-completion rate. Unconstrained chain-of-thought, even using strong reasoning LLMs such as \texttt{o1} and \texttt{o3-mini}, achieve near-zero success rates as the planner on constrained navigation, suggesting that scaling internal compute within a reactive paradigm does not substitute for explicit simulation of future states. All experiments were conducted between November and December 2024 using the best publicly available models and browser environments at that time, except for the o3-mini experiment which was performed in February 2025. While subsequent model and environment updates may yield higher absolute scores on these tasks, our results demonstrate that simulative reasoning provides a measurable and repeatable advantage over reactive policies under identical conditions. In the subsections below, we introduce the evaluation settings and discuss the results in more detail.

\subsection{Constrained Navigation}
\label{sec:flightqa}

\begin{figure*}[t]
    \centering
    \includegraphics[height=0.3\textwidth]{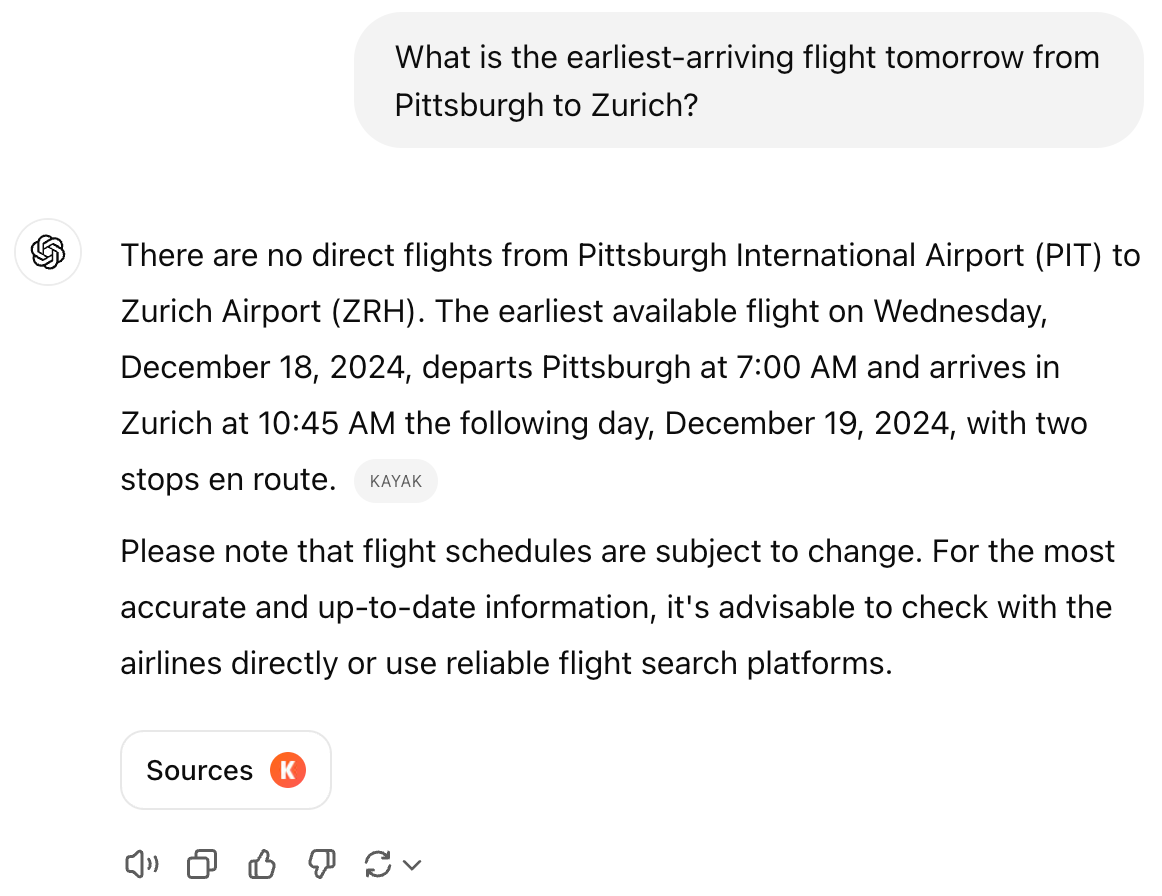}
    \hspace{0.05\textwidth}
    \includegraphics[height=0.3\textwidth]{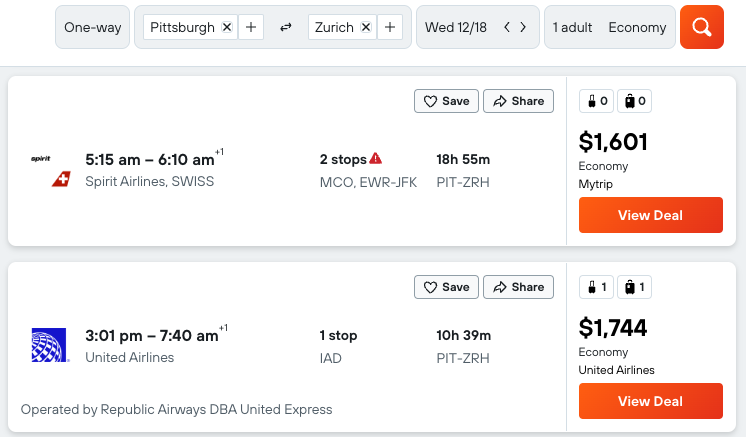}
    \caption{Faced with the question ``What is the earliest-arriving flight tomorrow from Pittsburgh to Zurich?'' ChatGPT-4o browsed the frontpage of Kayak.com and hallucinated a flight that arrives at 10:45am on the following day as the answer (\textbf{left}). Performing the search on Kayak.com, however, shows that the earliest-arriving flight lands in Zurich at 6:10am on the next day (\textbf{right}). The question was asked on December 17th, 2024.}
    \label{fig:chatgpt-kayak-example}
\end{figure*}

The first task category evaluates the agent's ability to navigate a complex, partially-observable state space under compositional constraints. This is a general challenge that arises whenever an agent must satisfy multiple simultaneous requirements while interacting with a complex interface, whether it be navigating a travel website, configuring a manufacturing system, or operating a scientific instrument.
We instantiate this category using flight search on live travel websites: the agent must execute a search query satisfying a list of constraints (e.g., one-way, from New York to Los Angeles, departing after 6 PM) and return a valid result. This demands multi-step reasoning and precise constraint handling in a dynamic, partially observable environment.
For many such questions, LLMs without environment grounding often hallucinate answers (see Figure~\ref{fig:chatgpt-kayak-example} for an example), underscoring the need for agents that can interact with live interfaces rather than relying on parametric knowledge alone.

\paragraph{Dataset} 
Existing benchmarks~\cite{he2024webvoyagerbuildingendtoendweb} have contributed valuable infrastructure for evaluating web agents in diverse real-world settings. However, these datasets are primarily constructed through self-instruct–based task generation followed by human verification, which, while ensuring breadth and realism, offers limited control over task structure and constraint complexity. Therefore, the resulting tasks may follow distributions favored by the language model used for generation, making it difficult to systematically assess how an agent’s reasoning performance scales with task difficulty or compositional complexity.
To address this limitation, we introduce FlightQA, a dataset designed for controlled and scalable evaluation of reasoning robustness in live web environments. 
We focus on flight search, a representative real-world use case that requires multi-step reasoning and precise constraint handling. 
In this setup, the user requests a flight satisfying a list of constraints (e.g., one-way, from New York to Los Angeles), and the agent must operate an online flight search interface to locate and return a valid option.
To systematically evaluate the agent's reasoning ability, we construct questions with varying number of constraints by iteratively adding to the list, which enables a type of \textit{counterfactual analysis} that controls for the confounding effect of specific constraint configurations. For instance, an agent capable of robust reasoning should 
succeed when an additional constraint is added to an existing query, whereas one relying on rote memorization will likely fail.

\begin{figure*}[t]
    \centering
    \includegraphics[width=0.9\textwidth]{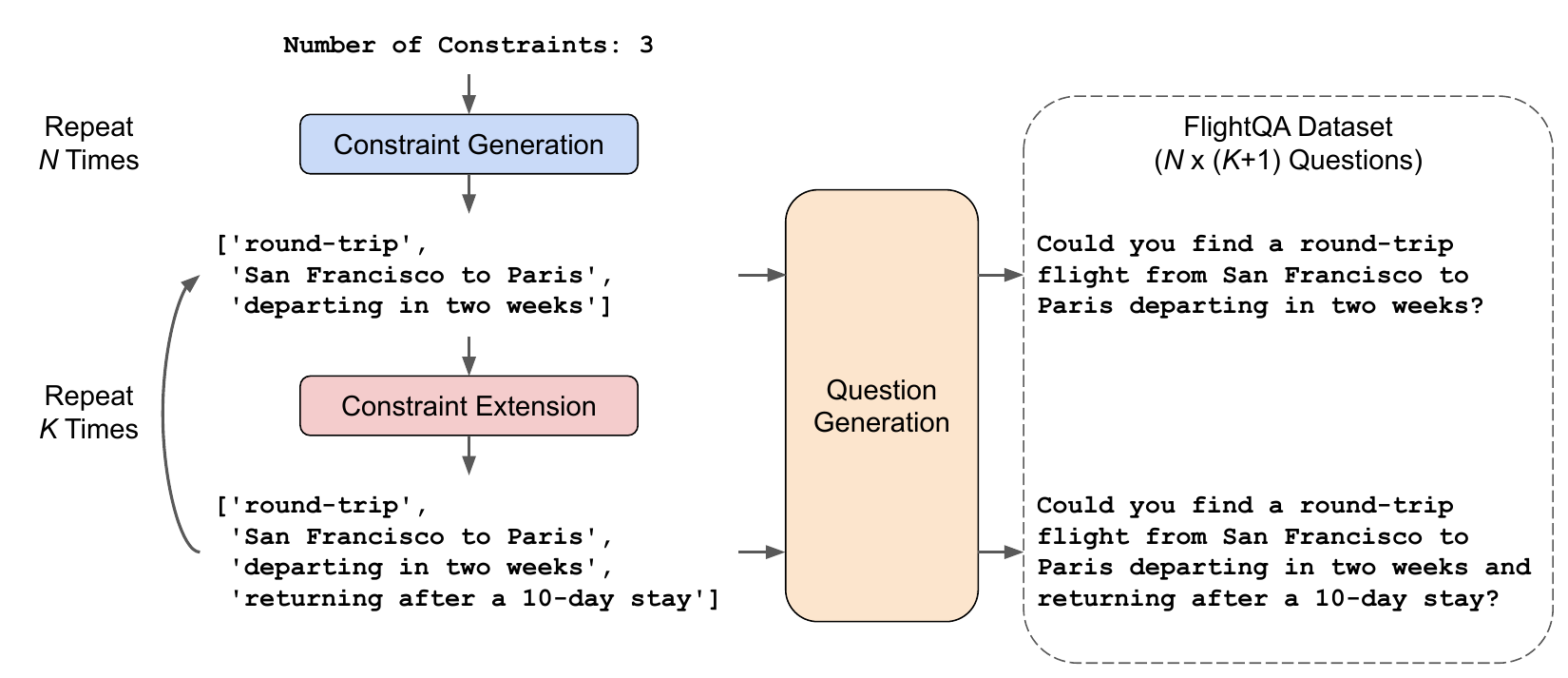}
    \caption{Illustration of the data generation process for the FlightQA dataset. We first prompt a LLM to generate $N$ lists of $C$ starting constraints (Constraint Generation). Then, we prompt the LLM to iteratively add constraints to the lists one by one, repeating for $K$ times (Constraint Extension). Finally, we prompt the LLM to convert each constraint list into a question in natural language (Question Generation).}
    \label{fig:fightqa-data-generation}
\end{figure*}

We illustrate our data collection process in Figure~\ref{fig:fightqa-data-generation}. To ensure scalability and controllability, we prompt a LLM to first generate a list of $C$ starting constraints, repeating for $N$ times. After that, we prompt the LLM to iteratively add constraints to the lists one at a time, repeating for $K$ times. Finally, we prompt the LLM to convert each constraint list into a question in natural language. In practice, we set $C = 3$, $N = 15$, and $K = 5$, which results in FlightQA, a dataset consisting of 90 questions with 15 sequences of constraint lists where the number of constraints increases from 3 to 8. We use \texttt{gpt-4o} to perform all the data generation steps. The initial question generation and question expansion prompts are included in Appendix~\ref{appendix:prompt-flightqa}.

\paragraph{Evaluation}
Because FlightQA involves querying live information from the open internet, it is impossible to establish ground truth answers due to the constantly evolving flight pricing and availability. Inspired by previous work on evaluation for generated text~\citep{deng2021compression}, we propose to evaluate the agent response based on two quality aspects: \textbf{groundedness} for whether the response is supported by the interaction history and \textbf{relevance} for whether the response satisfies user constraints to the extent allowed by the results (e.g., if the search results do not include any flight that satisfies all user constraints). 
Due to the strong ability of LLMs in evaluating generated text~\citep{liu-etal-2023-g}, we prompt LLMs to assess the two quality aspects of the agent response. Specifically, we include all browser observations in the agent's trajectory over $T$ steps $(o_1, o_2, … o_T)$, the constraint list, the question, and the agent response, and ask the LLM to provide judgment on the groundedness and relevance of the response. We further define an answer to be \textbf{correct} when it is both grounded and relevant. We also include the evaluation prompt in Appendix~\ref{appendix:prompt-flightqa}.

\paragraph{Experiment Setup}
We ran the experiments and evaluation using \texttt{gpt-4o} between November 24th, 2024 and December 9th, 2024. To evaluate whether LLMs trained with reinforcement learning to reason with unconstrained chain-of-thought can substitute for simulative reasoning, we also include an implementation of planner using \texttt{o1}~\cite{openai_learning_to_reason} and \texttt{o3-mini}~\cite{openai_o3_mini_2025} and ran them between February 3rd to 5th, 2025.  For the environment, we use BrowserGym~\citep{workarena2024}, a popular open-source browser sandbox. We stop each run when the agent provides a response or after the agent takes 30 actions, whichever comes first. We also mark the run as failed when the agent repeats the same action for 3 times consecutively or when the agent causes more than 3 errors while interacting with the browser.

\paragraph{Results}
We present the constrained navigation results in Table~\ref{tab:flightqa-results}. Compared to BrowsingAgent which fails completely in this task, simulative reasoning (System~II) improves the accuracy from 0\% to 32.2\%. Within our architecture, simulative reasoning shows superior performance over the reactive policy (System~I) with a 124\% improvement (significant at the 0.01 level). The structured state representation (e.g., observation summary and selective memory) also contributes to more coherent behavior, reducing the action error rate from 93.3\% in BrowsingAgent to 1.1\%. However, the reactive policy still results in frequent repetitions, which is mitigated by simulative reasoning (44.4\% $\rightarrow$ 18.9\%). Perhaps surprisingly, 
o1 and o3-mini achieve near-zero success rates when serving as planners with unconstrained chain-of-thought, suggesting that even strong reasoning models have limited success with planning without explicit simulation of future states. We do not include BrowsingAgent with o1 and o3-mini as the resulting agent frequently hallucinate answers without interacting with the webpage, which precludes them as viable agents. 
Those hallucinations, however, fool the LLM evaluator at significant rates, suggesting additional work is needed to secure its robustness.

\begin{table*}[t]
\centering
\setlength{\tabcolsep}{2pt}
{\renewcommand{\arraystretch}{1.2}
\small
\resizebox{\textwidth}{!}{%
\begin{tabular}{lrrr|rrrrr}
\toprule
 & \multicolumn{3}{l|}{\textbf{Performance (\%)}} & \multicolumn{5}{l}{\textbf{Outcomes (\%)}} \\
\textbf{Method} & \multicolumn{1}{l}{Correct} & \multicolumn{1}{l}{Grounded} & \multicolumn{1}{l|}{Relevant} & \multicolumn{1}{c}{\begin{tabular}[b]{@{}l@{}}Response \\ Returned\end{tabular}} & \multicolumn{1}{c}{\begin{tabular}[b]{@{}l@{}}Browser \\ Crashed\end{tabular}} & \multicolumn{1}{c}{\begin{tabular}[b]{@{}l@{}}Max Steps \\ Reached\end{tabular}} & \multicolumn{1}{c}{\begin{tabular}[b]{@{}l@{}}Repetitive \\ Actions\end{tabular}} & \multicolumn{1}{c}{\begin{tabular}[b]{@{}l@{}}Action \\ Errors\end{tabular}} \\
\midrule
OpenHands BrowsingAgent & 0.0 & 0.0 & 0.0 & 0.0 & 3.3 & 3.3 & 0.0 & 93.3 \\
\rowcolor{Gray}
\multicolumn{9}{l}{\textit{\modelname (Ours)}} \\
Unconstrained CoT (o1/o3-mini)\textsuperscript{\textdagger} & 1.1/3.3 & 1.1/4.4 & 1.1/3.3 & 1.1/4.4 & 11.1/3.3 & 40.0/51.1 & 37.8/32.2 & 10.0/8.9 \\
Reactive Policy (System~I) & 14.4 & 15.6 & 14.4 & 16.7 & 0.0 & 37.8 & 44.4 & 1.1 \\
Simulative Reasoning (System~II) & \textbf{32.2\textsuperscript{**}} & \textbf{36.7} & \textbf{32.2} & \textbf{38.9} & 1.1 & 40.0 & 18.9 & 1.1 \\ \bottomrule
\end{tabular}
}
}
\caption{Performance and outcome statistics for the constrained navigation task (FlightQA). Simulative reasoning (System~II) increases the accuracy from 0\% in OpenHands BrowsingAgent to 32.2\%, and outperforms the matched reactive policy (System~I) by 124\%. ** indicates being significantly higher than the second-best method at the statistical significance level of 0.01 ($p < 0.01$) based on pairwise t-test. \textsuperscript{\textdagger}We implement the planner with o1 and o3-mini, respectively.
}
\label{tab:flightqa-results}
\end{table*}

\begin{figure*}[t]
    \centering
    \includegraphics[width=0.8\textwidth]{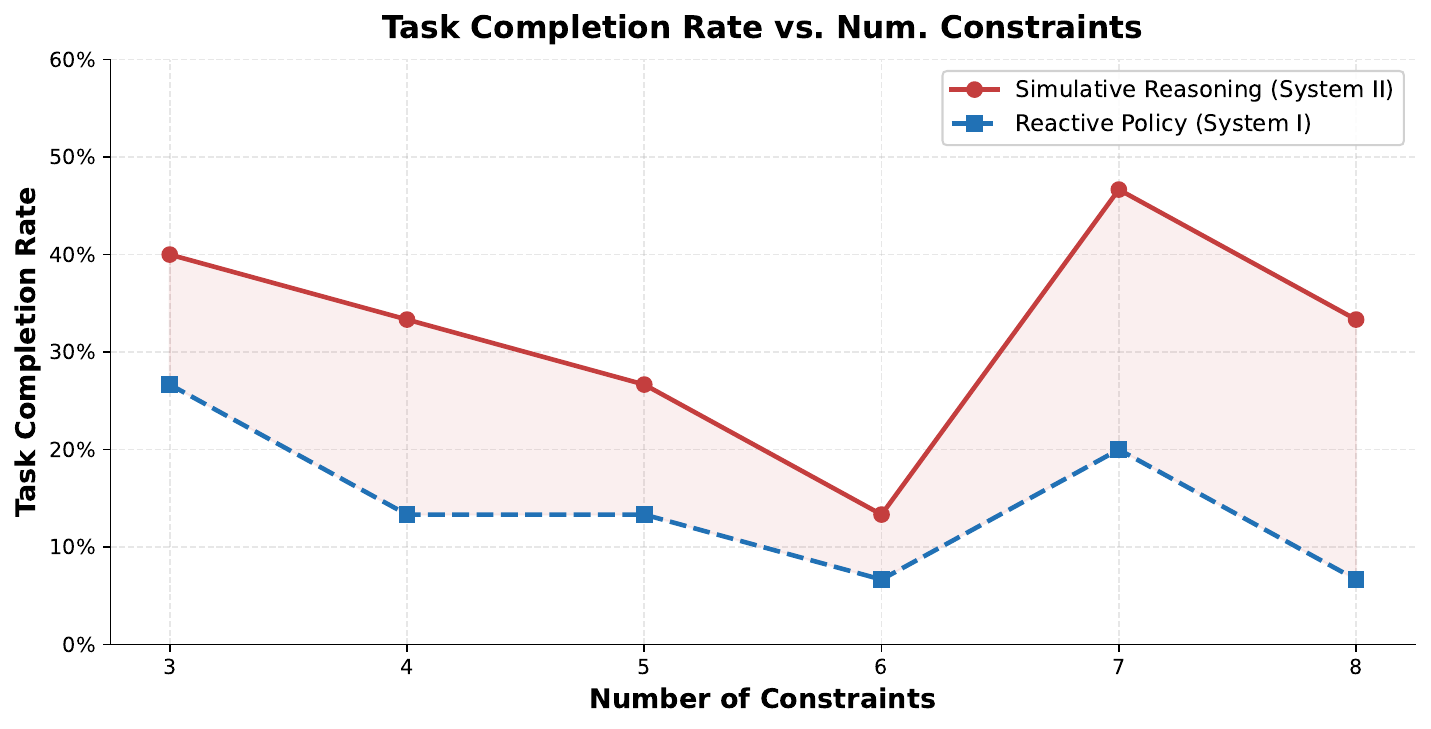}
    \caption{Completion rate as a function of the number of constraints in constrained navigation. The persistent gap between simulative reasoning (System~II) and the reactive policy (System~I) suggests the benefit stems from generalizable grounded counterfactual evaluation rather than task-specific tuning.}
    \label{fig:reasoning-analysis}
\end{figure*}

\paragraph{Analysis of Compositional Robustness} To evaluate how well simulative reasoning handles increasing compositional complexity, we visualize the percentage of correct responses vs. the number of constraints in Figure~\ref{fig:reasoning-analysis}. As the questions in FlightQA are generated based on iteratively expanded constraint lists, this analysis faithfully reflects the effect of increasing complexity while controlling for other confounders such as specific constraint sets. Simulative reasoning shows a consistent advantage over the reactive policy as the number of constraints increases, suggesting that the benefit stems from generalizable counterfactual evaluation rather than task-specific tuning. The performance for both methods decreases with more constraints initially but then increases sharply at 7 constraints before dropping again, which may reflect memorization in the backend LLM or implicit constraints in questions with fewer explicit constraints.

\subsection{Multi-Hop Information Aggregation}
 
The second task category evaluates the agent's ability to gather and synthesize information from multiple sources over long-horizon interactions. This challenge arises broadly whenever an agent must compose partial answers from multiple locations, whether it be querying several databases, consulting multiple documents, or surveying distributed web resources.
We instantiate this category using multi-hop, multi-website QA: given a question like ``\textit{What are the availabilities of the top-10 restaurants in Paris for a dinner next week?}'', the agent must first identify the relevant entities, then look up information about each across different sources, and finally compile the results. Whereas constrained navigation stresses the depth of interaction with individual interfaces, multi-hop information aggregation concerns the breadth of sources and the coherence of long-horizon reasoning.

\paragraph{Dataset} 
To evaluate agent abilities for multi-hop, multi-website QA, we adopt the FanOutQA~\citep{zhu-etal-2024-fanoutqa} dataset, which consists of questions of exactly this nature. Due to resource constraints, we evaluate on the first 100 examples of the dev set. As the results show, however, the smaller sample size is sufficient to show statistically significant differences between methods.  

\paragraph{Experiment Setup}
We ran the experiments using \texttt{gpt-4o-2024-05-13} between November 10th, 2024 and December 8th, 2024. We noticed that simulative reasoning deteriorates in performance when using the newer versions of \texttt{gpt-4o}, which may be due to additional training which changed the model's response patterns to the same prompts. 
We operate the browser using the same rules as in experiments for constrained navigation.

\paragraph{Results}
We present the multi-hop information aggregation results in Table~\ref{tab:fanoutqa-results}. Again, simulative reasoning (System~II) increases the accuracy from 17.0\% to 29.8\%, improving over the reactive policy (System~I) by 48.6\% (p-value = 0.011). BrowsingAgent achieves fair performance even though it cannot memorize information from different websites, often due to some questions in the dataset being answerable based on information from a single Wikipedia page (e.g., \textit{What are the publication dates for all of the Harry Potter books?}). Despite this, our architecture improves over BrowsingAgent even without simulative reasoning by dramatically reducing action errors (43\% $\rightarrow$ 10\%). Browser crashes make a sizable contribution to agent failures (24\% for our architecture), indicating room for improvement in the tooling for open-web navigation.

\begin{table}[t]
\centering
\setlength{\tabcolsep}{2pt}
{\renewcommand{\arraystretch}{1.2}
\small
\resizebox{\textwidth}{!}{%
\begin{tabular}{lrr|rrrrrr}
\toprule
 & \multicolumn{2}{l|}{\textbf{Performance (\%)}} & \multicolumn{6}{l}{\textbf{Outcomes (\%)}} \\
 \textbf{Method} & \multicolumn{1}{l}{Acc.} & \multicolumn{1}{l|}{Acc. (Strict)} & \multicolumn{1}{l}{\begin{tabular}[b]{@{}l@{}}Response \\ Returned\end{tabular}} & \multicolumn{1}{l}{\begin{tabular}[b]{@{}l@{}}Browser \\ Crashed\end{tabular}} & \multicolumn{1}{l}{\begin{tabular}[b]{@{}l@{}}Max Steps \\ Reached\end{tabular}} & \multicolumn{1}{l}{\begin{tabular}[b]{@{}l@{}}Repetitive \\ Actions\end{tabular}} & \multicolumn{1}{l}{\begin{tabular}[b]{@{}l@{}}Action \\ Error\end{tabular}} & \multicolumn{1}{l}{\begin{tabular}[b]{@{}l@{}}Parsing \\ Error\end{tabular}} \\ \midrule
OpenHands BrowsingAgent & 17.0 & 4.0 & 32.0 & 17.0 & 8.0 & 0.0 & 43.0 & 0.0 \\
\rowcolor{Gray}
\multicolumn{9}{l}{\textit{\modelname (Ours)}} \\
Reactive Policy (System~I) & 20.2 & 3.0 & 37.0 & 24.0 & 10.0 & 18.0 & 10.0 & 1.0 \\
Simulative Reasoning (System~II) & \textbf{29.8*} & 4.0 & \textbf{55.0} & 24.0 & 12.0 & 8.0 & 1.0 & 0.0 \\ \bottomrule
\end{tabular}
}
}
 
\caption{Performance and outcome statistics for multi-hop information aggregation (FanOutQA). Acc. (Strict) refers to the percentage of responses that exactly match the groundtruth. Simulative reasoning (System~II) increases the response rate and fact-level accuracy vs. the reactive policy (System~I) by 48.6\% and 47.5\%, respectively. * indicates being significantly higher than the second-best method at the 0.05 level based on pairwise t-test.}
 
\label{tab:fanoutqa-results}
\end{table}

\subsection{General Instruction Following}
 
The third task category evaluates the agent's ability to execute diverse natural-language instructions in a complex environment. This is the broadest category, encompassing a range of goals (e.g., information retrieval, content management, form submission) across varied interfaces, and tests whether the agent can flexibly adapt its behavior to different task specifications.
We instantiate this category using the WebArena benchmark~\citep{zhou2023webarena}: given instructions like ``\textit{Summarize customer reviews for Amazon Echo Dot 3rd generation},'' the agent must navigate appropriate interfaces to locate relevant content and carry out the task.

\paragraph{Dataset}
To evaluate general web automation capabilities, we adopt the WebArena~\citep{zhou2023webarena} benchmark, a standard environment for testing web agents which features a range of simulated websites including a Reddit-like social forum, a shopping site, a GitLab-based code management platform, a map, and a Wikipedia-like encyclopedia. Following the evaluation for Multi-Hop, Multi-Website QA, we take a random subset of 100 examples.

\paragraph{Experiment Setup}
We run the experiments using \texttt{gpt-4o} over BrowserGym accessed via the OpenHands platform which provides a uniform evaluation procedure. Because WebArena demands a specific response format for evaluation, we rewrote the agent description to steer the agent answer format accordingly (Appendix~\ref{appendix:webarena-adaptation}). We keep all other environment rules the same as previous experiments, except for setting the maximum allowed steps to 15 which is consistent with the default setting of WebArena. 

\paragraph{Results}
We present the general instruction following results in Table~\ref{tab:webarena-results}. Continuing the pattern from the previous task categories, simulative reasoning (System~II) improves over the reactive policy (System~I) by 21.1\%, and over BrowsingAgent by up to 91.7\%. Due to following the environment and evaluator provided by OpenHands, which prioritizes open-web browsing and differs from the standard benchmark setup, the success rates are not directly comparable to those reported in prior work~\cite{zhou2023webarena}. Nevertheless, these results demonstrate that the relative advantage of simulative reasoning persists across a third, qualitatively distinct task category.

\begin{table}[t]
\centering
\setlength{\tabcolsep}{2pt}
{\renewcommand{\arraystretch}{1.2}
\small
\begin{tabular}{lr}
\toprule
\textbf{Method} & \multicolumn{1}{l}{\textbf{Success Rate (\%)}} \\ \midrule
OpenHands BrowsingAgent & 12.0 \\
\rowcolor{Gray}
\multicolumn{2}{l}{\textit{\modelname (Ours)}} \\
Reactive Policy (System~I) & 19.0 \\
Simulative Reasoning (System~II) & \textbf{23.0} \\
\bottomrule
\end{tabular}
}
 
\caption{Results on general instruction following (WebArena, random 100-sample subset). Simulative reasoning (System~II) improves over the reactive policy (System~I) by 21.1\% and over BrowsingAgent by 91.7\%.
}
\label{tab:webarena-results}
\end{table}

\section{Limitations}
Due to the modular pipeline and thorough exploration of multiple candidate plans, simulative reasoning incurs higher computational cost than reactive policies at inference time. Speeding up world-model-based reasoning with appropriate caching, abstention, and parallelization strategies is an important part of our future work. In addition, agent capabilities can be limited by the tooling. For example, with open-source browser environments, web agents are often blocked by Captcha or anti-scraping tools from certain websites. Deeper integration with user browser can help solve this issue. As agent-based automation become more integrated into browsing and computer-use workflows, we also encourage conversations around fair use and protocols around agent access of certain websites. We are currently only using the text portion of the webpage observations, which can miss information like images and layout information (e.g., occlusions). While existing work are experimenting with visual-based web browsing, it is still challenging to combine multimodal perception and planning, which we are excited to keep working on.

\section{Conclusion}
In this paper, we have argued that simulative reasoning through a world model (System~II) provides a general-purpose planning mechanism for agentic systems, improving upon reactive policies (System~I) by grounding decisions in predicted future states rather than pattern-matched responses.
We introduced \modelname, an architecture that instantiates this principle using an LLM-based world model with natural-language belief states, and evaluated it across three qualitatively distinct task categories: constrained navigation, multi-hop information aggregation, and general instruction following.
Across all categories, simulative reasoning consistently outperforms the matched reactive baseline, with the advantage persisting as task complexity increases. 
These results suggest that the benefit stems from a generalizable property of the reasoning pattern (i.e., grounded counterfactual evaluation) rather than domain-specific tuning, providing initial empirical support for the hypothesis that simulative reasoning is a core component of general-purpose agentic planning.
 
Looking ahead, the principle of simulative reasoning extends naturally beyond the web-browser environment studied here to embodied, social, and physical domains, wherever an agent must anticipate the consequences of its actions before committing.
On the capability side, improving world model accuracy and exploring how to regulate when simulation is worth its computational cost are important directions.
On the learning side, we see substantial potential in leveraging world models as simulators for continual reinforcement learning: combining learning within a world-model sandbox with periodic grounding on real data can improve sample efficiency and robustness, especially when access to real environments is constrained by resources, safety, or legal considerations.
On the safety and alignment side, we look forward to engaging the community in discussions about how explicit world-model reasoning may help build agents that remain transparent and aligned with human values.

\section*{Acknowledgment}
This work was supported in part by the Samsung GRO Project ``Efficient Designs for Generative and Agent LLM Development.'' We thank Graham Neubig and Zora Wang from NeuLab; Yilin Shen and Hongxia Jin from Samsung Research; Zhoujun Cheng, Shibo Hao, and Xinyu Pi from MixLab; Han Guo, Nicholas Ho, and Bowen Tan from SAILING Lab; Li Erran Li from AWS, and Sarah Cheah and Hector Ren from MBZUAI for their insightful feedback and discussions. We are also grateful for their helpful suggestions throughout the project. Any opinions, findings, and conclusions or recommendations expressed in this material are those of the authors and do not necessarily reflect the views of Samsung.

\bibliographystyle{plain} 
\bibliography{ref} 

\newpage

\appendix

\section{\modelname Inference Pseudo-Code}
The pseudo-code below outlines \modelname’s planning routine at timestep $t$; procedures like replanning decision and  are omitted for simplicity.
\label{appendix:sira-inference}
\begin{algorithm}[h]
\caption{Inference algorithm of \modelname for agentic planning at time $t$}
\begin{algorithmic}[1]
\REQUIRE Observation $o_t$, Goal $g$
\REQUIRE Encoder $h$, Proposer policy $\tilde{\pi}$, World model $f$, Critic $v$, Actor $\alpha$
\REQUIRE Planning horizon $T'$
\vspace{0.5em}

\STATE \textbf{(Perception)}
\STATE Infer current belief state: $\hat{s}_{t} \leftarrow \argmax_{\hat{s}} p_h(\hat{s} \mid o_t)$

\vspace{0.5em}
\STATE \textbf{(Planning via Simulation)}
\STATE Initialize best score $V^\star \leftarrow -\infty$, best simulated action sequence $A'^\star \leftarrow \varnothing$
\STATE Initialize a search tree with root node $(\hat{s}_t)$

\WHILE{DFS budget not exhausted}
    \STATE Initialize rollout variables: $\hat{s}_t^{(0)} \leftarrow \hat{s}_t,\;\; k \leftarrow 1,\;\; \mathcal{A}' \leftarrow [\ ]$

    \WHILE{$k < T'$}
        \STATE Propose a high-level simulated action:
        $a'_{t+k} \leftarrow \argmax_{a'} p_{\tilde{\pi}}(a' \mid \hat{s}_{t+k}^{(k)})$
        \STATE Append $a'_{t+k}$ to $\mathcal{A}'$
        \STATE Predict next belief state using the world model:
        $\hat{s}_{t+k+1}^{(k+1)} \leftarrow \argmax_{\hat{s}} p_f(\hat{s} \mid \hat{s}_{t+k}^{(k)}, a'_{t+k})$
        \STATE $k \leftarrow k+1$
    \ENDWHILE

    \STATE Evaluate terminal belief state at depth $T'$ using critic:
    $V \leftarrow v\!\left(\hat{s}_{t+T'}^{(T')}\right)$
    \COMMENT{$v(\cdot)$ reflects cumulative goal progress $\sum_{i=t}^{t+T'-1} \gamma_i r(g,\hat{s}_i) + \gamma_{t+T'} V_{\pi,f}^g(\hat{s}_{t+T'})$}

    \IF{$V > V^\star$}
        \STATE $V^\star \leftarrow V$
        \STATE $A'^\star \leftarrow \mathcal{A}'$ \COMMENT{store best simulated action sequence}
    \ENDIF
\ENDWHILE

\vspace{0.5em}
\STATE \textbf{(Acting)}
\STATE Infer executable next step action in the concrete action space:
$a_t \leftarrow \argmax_{a} p_\alpha(a \mid \hat{s}_t, A'^*)$

\vspace{0.5em}
\STATE \textbf{(Execute in Environment)}
\STATE Issue $a_t$ to the environment, observe next $o_{t+1}$

\vspace{0.5em}
\STATE \textbf{return} $a_t$
\end{algorithmic}
\end{algorithm}
\raggedbottom

\section{Details on Web Browsing Environment}
\label{appendix:browsing-environment}
\vspace{-6pt}
\begin{tcolorbox}[width=\textwidth,colback={blue!30!},title={Example Observation},colbacktitle=black,coltitle=white,boxrule=1pt]
{\scriptsize
\begin{verbatim}
URL https://www.google.com/travel/flights
Scroll Position: 0, Window Height: 720, Webpage Height: 3024, Remaining Pixels: 2304, Scrolling 
Progress: 23.8%
RootWebArea 'Google Flights - Find Cheap Flight Options & Track Prices'
    [149] banner ''
        [160] button 'Main menu', clickable, expanded=False
            [161] image '' 
        [168] link 'Google', clickable
        StaticText 'Skip to main content'
        StaticText 'Accessibility feedback'
        [186] navigation ''
            [189] link 'Travel'
                [193] image ''
            [197] link 'Explore'
                [201] image ''
            [205] link 'Flights'
                [209] image ''
            [213] link 'Hotels'
                [217] image ''
            [221] link 'Vacation rentals'
                [225] image ''
        [235] button 'Change appearance', hasPopup='menu', expanded=False
            [240] image ''
        [249] button 'Google apps', clickable, expanded=False
            [250] image ''
        [251] link 'Sign in', clickable
    [342] image ''
    StaticText 'Flights'
    [346] search 'Flight'
        [355] combobox 'Change ticket type. \u200bRound trip', live='polite', relevant='additions 
              text', hasPopup='listbox', expanded=False, controls='i9'
            [364] image ''
        [399] button '1 passenger, change number of passengers.', hasPopup='dialog'
            [404] image ''
            [406] image ''
        [522] combobox 'Change seating class. \u200bEconomy', live='polite', relevant='additions 
              text', hasPopup='listbox', expanded=False, controls='i22'
                [529] image ''
        [576] combobox 'Where from?' value='Pittsburgh', clickable, autocomplete='inline', 
              hasPopup='menu', expanded=False
        [580] image ''
        [628] button 'Swap origin and destination.', disabled=True
                [631] image ''
        [638] combobox 'Where to?', clickable, focused, autocomplete='inline', hasPopup='menu', 
              expanded=False
        [641] image ''
        generic '', hidden=True
        [690] image ''
        [691] textbox 'Departure', clickable, describedby='i32'
        [712] textbox 'Return', clickable, describedby='i32'
        generic '', hidden=True
        [857] button 'Explore destinations'
    [866] Section ''
        [867] heading 'Find cheap flights from Pittsburgh to anywhereMore information on suggested 
              flights.'
            [871] button 'More information on suggested flights.', hasPopup='menu'
                [873] image ''
        [904] list '', clickable
            [905] listitem ''
                StaticText 'Pittsburgh'
            [907] listitem ''
                [908] button 'Cleveland'
            [909] listitem ''
                [910] button 'Columbus'
            [911] listitem ''
                [912] button 'Akron'
    StaticText 'San Francisco'
    StaticText '\$128'
    StaticText 'Jan 9 — Jan 16'
    StaticText '1 stop'
    StaticText '·'
    StaticText '10 hr 30 min'
    StaticText 'New York'
    StaticText '\$68'
    StaticText 'Dec 7 — Dec 14'
\end{verbatim}
}
\end{tcolorbox}

\section{Prompts for Web Browsing Implementation}
\label{appendix:prompts-browsing-implementation}

\begin{tcolorbox}[width=\textwidth,colback={blue!30!},title={Prompt for Agent Identity},colbacktitle=black,coltitle=white,boxrule=1pt]    
\# Name:
\newline
Web Browsing Agent
\newline
\newline
\# Description:
\newline
An information and automation assistant who responds to \
user instructions by browsing the internet. The assistant strives to answer each question 
accurately, thoroughly, efficiently, and politely, and to be forthright when it is 
impossible to answer the question or carry out the instruction. The assistant will 
end the task once it sends a message to the user.
\newline \newline
\# Observation Space:
\newline
The text representation and screenshot of the part of webpage visible in the viewport of a browser. \
Here is an abstract description of the information available in the webpage text representation:
\newline \newline
- Identification Information:
\begin{itemize}
    \item[-] URL: The web address that specifies the location of the webpage.
    \item[-] Document Properties: Attributes such as scroll position and viewport dimensions that describe the current viewing context.
\end{itemize}
- Structural Hierarchy:
\begin{itemize}
    \item[-] Root Element: The primary container for the webpage, indicating its overall theme or purpose.
    \item[-] Nested Elements: A hierarchy of sections, containers, and components that organize content logically (e.g., headers, footers, sidebars).
\end{itemize}
- Interactive Components:
\begin{itemize}
    \item[-] Links: Elements that can be clicked to navigate to other pages or sections, often labeled descriptively.
    \item[-] Buttons: Interactive controls that trigger actions (e.g., submitting forms, opening menus).
\end{itemize}
- Content Types:
\begin{itemize}
    \item[-] Text: Main content, headings, and subheadings that provide information and context.
    \item[-] Images and Media: Visual elements that enhance the understanding or appeal of the content.
    \item[-] Forms and Inputs: Fields for user input, including text boxes, dropdowns, and checkboxes.
\end{itemize}
- Functional Areas:
\begin{itemize}
    \item[-] Navigation Menus: Organized sets of links that allow users to explore different sections of the site.
    \item[-] Search Interface: Components that enable users to search for content within the site, including input fields and associated buttons.
\end{itemize}
- State Information:
\begin{itemize}
    \item[-] Visibility and Expand/Collapse States: Indicators showing whether certain elements are active, visible, or in a collapsed state, impacting user interaction.
    \item[-] Focus States: Information on which elements are currently focused, important for keyboard navigation and accessibility.
\end{itemize}
\end{tcolorbox}  

\newpage

\begin{tcolorbox}[width=\textwidth,colback={blue!30!},title={Prompt for Agent Identity (Continued)},colbacktitle=black,coltitle=white,boxrule=1pt]
- Accessibility Features:
\begin{itemize}
    \item[-] Role and Description Information: Metadata that provides context about the purpose of elements, useful for screen readers and assistive technologies.
\end{itemize}
- General User Considerations:
\begin{itemize}
    \item[-] Navigation: Recognizing how to traverse the webpage using links and buttons.
    \item[-] Interactivity: Understanding how to engage with forms, search fields, and dynamic components.
    \item[-] Content Engagement: Identifying and interpreting various content types to glean necessary information.
\end{itemize}
~\\
\# Action Space:
\newline
13 different types of actions are available.
\newline \newline
noop(wait\_ms: float = 1000)
\begin{itemize}
    \item[] Examples:
    \begin{itemize}
        \item[] noop()
        \item[] noop(500)
    \end{itemize}
\end{itemize}

send\_msg\_to\_user(text: str)
\begin{itemize}
    \item[] Examples:
    \begin{itemize}
        \item[] send\_msg\_to\_user('Based on the results of my search, the city was built in 1751.')
    \end{itemize}
\end{itemize}

scroll(delta\_x: float, delta\_y: float)
\begin{itemize}
    \item[] Examples:
    \begin{itemize}
        \item[] scroll(0, 200)
        \item[] scroll(-50.2, -100.5)
    \end{itemize}
\end{itemize}

fill(bid: str, value: str)
\begin{itemize}
    \item[] Examples:
    \begin{itemize}
        \item[] fill('237', 'example value')
        \item[] fill('45', 'multi-line\textbackslash nexample')
        \item[] fill('a12', 'example with "quotes"')
    \end{itemize}
\end{itemize}

select\_option(bid: str, options: str | list[str])
\begin{itemize}
    \item[] Examples:
    \begin{itemize}
        \item[] select\_option('a48', 'blue')
        \item[] select\_option('c48', ['red', 'green', 'blue'])
    \end{itemize}
\end{itemize}

click(bid: str, button: Literal['left', 'middle', 'right'] = 'left', modifiers: list[typing.Literal['Alt', 'Control', 'Meta', 'Shift']] = [])
\begin{itemize}
    \item[] Examples:
    \begin{itemize}
        \item[] click('a51')
        \item[] click('b22', button='right')
        \item[] click('48', button='middle', modifiers=['Shift'])
    \end{itemize}
\end{itemize}

dblclick(bid: str, button: Literal['left', 'middle', 'right'] = 'left', modifiers: list[typing.Literal['Alt', 'Control', 'Meta', 'Shift']] = [])
\begin{itemize}
    \item[] Examples:
    \begin{itemize}
        \item[] dblclick('12')
        \item[] dblclick('ca42', button='right')
        \item[] dblclick('178', button='middle', modifiers=['Shift'])
    \end{itemize}
\end{itemize}
\end{tcolorbox}  

\begin{tcolorbox}[width=\textwidth,colback={blue!30!},title={Prompt for Agent Identity (Continued)},colbacktitle=black,coltitle=white,boxrule=1pt]    
hover(bid: str)
\begin{itemize}
    \item[] Examples:
    \begin{itemize}
        \item[] hover('b8')
    \end{itemize}
\end{itemize}

press(bid: str, key\_comb: str)
\begin{itemize}
    \item[] Examples:
    \begin{itemize}
        \item[] press('88', 'Backspace')
        \item[] press('a26', 'Control+a')
        \item[] press('a61', 'Meta+Shift+t')
    \end{itemize}
\end{itemize}

focus(bid: str)
\begin{itemize}
    \item[] Examples:
    \begin{itemize}
        \item[] focus('b455')
    \end{itemize}
\end{itemize}

clear(bid: str)
\begin{itemize}
    \item[] Examples:
    \begin{itemize}
        \item[] clear('996')
    \end{itemize}
\end{itemize}

drag\_and\_drop(from\_bid: str, to\_bid: str)
\begin{itemize}
    \item[] Examples:
    \begin{itemize}
        \item[] drag\_and\_drop('56', '498')
    \end{itemize}
\end{itemize}

upload\_file(bid: str, file: str | list[str])
\begin{itemize}
    \item[] Examples:
    \begin{itemize}
        \item[] upload\_file('572', 'my\_receipt.pdf')
        \item[] upload\_file('63', ['/home/bob/Documents/image.jpg', '/home/bob/Documents/file.zip'])
    \end{itemize}
\end{itemize}

Only a single action can be provided at once. Example:
fill('a12', 'example with "quotes"')
\newline \newline
\# Instruction:
\{user\_instruction\}
\newline \newline
\# Current Date and Time:
\{current\_datetime\}
\end{tcolorbox}  

\begin{tcolorbox}[width=\textwidth,colback={blue!30!},title={Prompt for Encoder},colbacktitle=black,coltitle=white,boxrule=1pt]    
\# Observation:
\newline
\{observation\}
\newline
\newline
\# State:
\newline
Describe all the elements in the current webpage observation. Note any dialog, progress indicators, or error messages. Include any interactive elements and their values or if they are blank. Note any detailed information such as facts, entities, or data that are relevant 
to the task. Report any error messages like whether the last action was correct. 
Try to be as comprehensive and detailed as possible.
\newline
\newline
Wrap your response in the tag <state> and </state>.
\end{tcolorbox}  

\newpage

\begin{tcolorbox}[width=\textwidth,colback={blue!30!},title={Prompt for Policy},colbacktitle=black,coltitle=white,boxrule=1pt]    
\{memory\}
\newline
\newline
\# Current State:
\newline
\{state\}
\newline
\newline
\# Intent:
\newline
Describe the action the assistant should take next to carry out the user's 
instruction. 
Avoid using phrases such as "To accomplish the goal," "I will," "To 
proceed.". Avoid ending with phrases like "to execute the search." 
Describe one action at a time and avoid combining multiple steps. 
Refrain from mentioning specific element IDs as they may change 
during execution. Limit your response to one phrase and include any details 
that help select the correct action. Be creative and propose novel 
methods to achieve the goal. Avoid creating accounts without user 
permission or providing personal information. Concrete example 
would be "Go to the home page of Google Flights." and "Click on the 'Search' button."
\newline
\newline
Wrap your response in the following format:
\newline
\newline
<think>
\newline
Your thoughts and reasoning process
\newline
</think>
\newline
\newline
<intent>
\newline
Description of the action to perform next
\newline
</intent>
\end{tcolorbox}

\begin{tcolorbox}[width=\textwidth,colback={blue!30!},title={Prompt for World Model},colbacktitle=black,coltitle=white,boxrule=1pt]    
\{memory\}
\newline
\newline
\# Current State:
\newline
\{state\}
\newline
\newline
\# Memory Update:
\newline
\{memory\_update\}
\newline
\newline
\# Action Intent:
\newline
\{plan\}
\newline
\newline
\# Next State:
\newline
Describe all the elements in the webpage after the agent attempts to carry out the intent. 
Note that the execution may not be successful, so you will have to infer the result of the action. 
Note any dialog, progress indicators, or error messages. Include any interactive elements and their 
values or if they are blank. Note any detailed information such as facts, entities, or data that are relevant 
to the task. Report any error messages displayed. Try to be as comprehensive and detailed as possible.
\newline
\newline
Wrap your response in the following format:
\newline
\newline
<next\_state>
\newline
Follow the format of the current state description. Use present tense. 
Avoid starting phrases like "Based on the interaction history, current state, and current intent".
\newline
</next\_state>
\end{tcolorbox} 

\newpage

\begin{tcolorbox}[width=\textwidth,colback={blue!30!},title={Prompt for Critic},colbacktitle=black,coltitle=white,boxrule=1pt]    
\{memory\}
\newline
\newline
\# Final State:
\newline
\{state\}
\newline
\newline
\# Task Success and Progress:
\newline
Your task is to evaluate the performance of the agent. Given the agent's instruction, interaction history, the final 
state of the webpage, and the agent's responses to the user if any, your goal is to decide whether the agent's execution 
is successful or not. If the current state is a failure but it looks like the agent is on the right track towards 
success, you should also output as such.
\newline
\newline
Wrap your response in the following format:
\newline
\newline
<think>
\newline
Your thoughts and reasoning process
\newline
</think>
\newline
\newline
<status>
\newline
"success" or "failure"
\newline
</status>
\newline
\newline
<on\_the\_right\_track>
\newline
"yes" or "no"
\newline
</on\_the\_right\_track>
\end{tcolorbox} 

\begin{tcolorbox}[width=\textwidth,colback={blue!30!},title={Prompt for Memory Update},colbacktitle=black,coltitle=white,boxrule=1pt]    
\{memory\}
\newline
\newline
\# State:
\newline
\{state\}
\newline
\newline
\# Action Intent:
\newline
\{plan\}
\newline
\newline
\# Memory Update:
\newline
Summarize the changes in the webpage observation that should be remembered for 
achieving your goal and for predicting the next state. Note any new elements, 
any elements no longer visible, or any changes in the content of existing elements. 
Also note if there is no change. Include any other inferred information that may help 
you decide the next action, such as whether an action intent is successful, or whether 
progress has been made or reversed. Do not include your next planned actions. Revise 
your belief from previous history if the current state contradicts it.
\newline
\newline
Wrap your response in the tag <memory\_update> and </memory\_update>.
\end{tcolorbox} 

\newpage

\begin{tcolorbox}[width=\textwidth,colback={blue!30!},title={Prompt for Actor},colbacktitle=black,coltitle=white,boxrule=1pt]    
\{memory\}
\newline
\newline
\# Observation:
\newline
\{observation\}
\newline
\newline
\# Current State:
\newline
\{state\}
\newline
\newline
\# Current Intent:
\newline
\{plan\}
\newline
\newline
\# Action:
\newline
Choose an API call that will carry out the intent when executed in the webpage. 
Use only one action at a time. You must not enclose bid inputs in [brackets] but instead in 'single quotes'. 
Interact only with elements in the current step observation. Your response 
will be executed as a Python function call, so ensure it adheres to the format 
and argument data type specifications defined in the action space.
\newline
\newline
Wrap your response in the tag <action> and </action>.
\end{tcolorbox} 

\newpage

\begin{tcolorbox}[width=\textwidth,colback={blue!30!},title={Prompt for Action Clustering},colbacktitle=black,coltitle=white,boxrule=1pt]    
Here is the action space for a browser agent to navigate in a webpage:
\newline
\newline
16 different types of actions are available:
\newline
\newline
noop(wait\_ms: float = 1000)
\newline
\newline
send\_msg\_to\_user(text: str)
\newline
\newline
scroll(delta\_x: float, delta\_y: float)
\newline
\newline
fill(bid: str, value: str)
\newline
\newline
select\_option(bid: str, options: str | list[str])
\newline
\newline
click(bid: str, button: Literal['left', 'middle', 'right'] = 'left', modifiers: list[typing.Literal['Alt', 'Control', 'Meta', 'Shift']] = [])
\newline
\newline
dblclick(bid: str, button: Literal['left', 'middle', 'right'] = 'left', modifiers: list[typing.Literal['Alt', 'Control', 'Meta', 'Shift']] = [])
\newline
\newline
hover(bid: str)
\newline
\newline
press(bid: str, key\_comb: str)
\newline
\newline
focus(bid: str)
\newline
\newline
clear(bid: str)
\newline
\newline
drag\_and\_drop(from\_bid: str, to\_bid: str)
\newline
\newline
upload\_file(bid: str, file: str | list[str])
\newline
\newline
go\_back()
\newline
\newline
go\_forward()
\newline
\newline
goto(url: str)
\newline
\newline
Only a single action can be provided at once. Example:
    fill('a12', 'example with "quotes"')
\newline \newline
Below, you will find lists of intents, or natural language descriptions of actions that, when executed, will translate to one of the function calls above. 
The intents will be provided in the following JSON format:

\begin{verbatim}
```json
{
  "intent_id": "intent description"
}
```
\end{verbatim}

Your task is to cluster list of intents into semantically equivalent groups, where each group represents intents that lead to the same action when executed 
(i.e., navigating to the Google homepage is translated to goto('https://www.google.com')) and would therefore correspond to the same API call 
in a Playwright browser. Intents that use different wording but convey the same action should be grouped together. Try to minimize the number of clusters.
\end{tcolorbox}

\begin{tcolorbox}[width=\textwidth,colback={blue!30!},title={Prompt for Action Clustering (Continued)},colbacktitle=black,coltitle=white,boxrule=1pt]   
Represent the clustering results using a JSON object where each cluster has a unique identifier, and each identifier maps to a list of actions in that cluster. 
See below for an abstract example:

\begin{verbatim}
```json
{
  "cluster_id": {
    "intent": "representative intent name for this cluster",
    "candidates": [
      "<list of intent ids that belong to this cluster>
    ]
  }
}
```
\end{verbatim}

Concrete Example 1:
\newline \newline
Dictionary of Intents:

\begin{verbatim}
```json
{
  "0": "Navigate to the Google homepage by entering its URL.",
  "1": "Go to the Google homepage.",
  "2": "Go to the Google homepage",
  "3": "Go to the Google homepage by navigating to
        'https://www.google.com'",
  "4": "Go to the home page of Google"
}
```
\end{verbatim}

["Navigate to the Google homepage by entering its URL.", "Go to the Google homepage.", "Go to the Google homepage", "Go to the Google homepage by navigating to \"https://www.google.com\"", "Go to the home page of Google"]
\newline \newline
Clustering Results:

\begin{verbatim}
```json
{
  "cluster_1": {
    "intent": "Navigate to the Google homepage",
    "candidates": [0, 1, 2, 3, 4]
  }
}
```
\end{verbatim}

Concrete Example 2:
\newline \newline
Dictionary of Intents:
\newline \newline
\{action\_candidate\_json\}
\newline \newline
Clustering Results:
\end{tcolorbox}

\subsection{Adaptation for WebArena Evaluation}
\label{appendix:webarena-adaptation}

\begin{tcolorbox}[width=\textwidth,colback={blue!30!},title={Agent Description for WebArena Evaluation},colbacktitle=black,coltitle=white,boxrule=1pt]
An information and automation assistant that interacts with the browser 
and responds to user instructions. The response follows the following rules: 
1. When the intent is a question, and a complete answer to the question has been found, 
then send the answer to the user; 2. the intent wants to locate specific information or navigate to 
a particular section of a site, and the current page satisfies, then stop and tell the user you found the required information; 
3. the intent want to conduct an operation, and has been done, then stop and tell the user the operation has been completed.

The assistant should try to achieve the goal in the current site without navigating to sites 
like Google. Be forthright when it is impossible to answer the question or carry out the task. 
The assistant will end the task once it sends a message to the user.
\end{tcolorbox}

\newpage

\section{Prompts for Generating and Evaluating on the FlightQA Dataset}
\label{appendix:prompt-flightqa}

\begin{tcolorbox}[width=\textwidth,colback={blue!30!},title={Prompt for Generating Initial Constraints and Questions},colbacktitle=black,coltitle=white,boxrule=1pt] 
\textbf{System}:

You are a creative writer who is an expert at crafting questions to help train assistants who answer user queries. Current date and time: \{current\_datetime\}
\\~\\
\textbf{Instruction}:

Your task is to create a robust benchmark for evaluating an AI's ability to search for flights through a platform like Google Flights.
To ensure the dataset effectively represents real-world use cases. Here are some important factors to consider:
\\~\\
1. Diversity of Queries

- Range of Destinations: Include both common and obscure destinations to test how well the model handles varying levels of demand.

- Dates and Duration: Include different date ranges, including last-minute flights, peak travel dates (like holidays), and extended trips. Ensure there's a variety in trip duration as well.

- Passenger Variability: Include solo travelers, families, and group travel (e.g., one adult vs. two adults and two children) since these change the search parameters and price results.

- Class and Preference: Vary preferences like cabin class (economy, business, first) and filters (non-stop, one stop, preferred airlines, etc.).

- Budget Constraints: Test price sensitivity by setting different budget limits to see how well the AI handles trade-offs.
\\~\\
2. Complexity of Requirements

- Multi-Leg Flights: Add queries for multi-city trips or those requiring complex layovers.

- Dynamic Constraints: Include queries with dynamic constraints, such as ``find the cheapest flight but depart between 8-10 AM,'' to see if the model can adapt its search within specific time frames.

- Conditional Preferences: Test cases where users might want options based on multiple conditions, like ``either find the cheapest non-stop or the shortest two-stop option.''
\\~\\
In practice, the questions typically vary in the following dimensions: 

- Ticket type (round-trip, one-way, etc.)

- Routes (origin and destination)

- Layover location(s)

- Dates (departure and/or return)

- Flight time (departure and arrival)

- Total flight time

- Airlines

- Cabin class (economy, business, etc.)

- Aircraft

- Eco-friendly options (CO2 Emissions)
\\~\\
Given a number of constraints, 
you should first provide a list of constraints, with the number of constraints equal to the specification. 
After that, you will generate a question a typical user will ask which imposes those constraints. 
You should repeat this for at least 7 times to generate a set of questions with simple language. 
Make sure that the number of constraints in the question matches the number of constraints specified.
\\~\\
Do not include constraints about the number of passengers. 
If the constraint is a date, you can use relative dates (e.g., "tomorrow", "next month", "after 8 PM", etc.). 
Avoid using specific dates like "December 25th" to ensure the questions are relevant throughout the year.
\\~\\
Your response should follow the JSON format below: 
\end{tcolorbox}

\begin{tcolorbox}[width=\textwidth,colback={blue!30!},title={Prompt for Generating Initial Constraints and Questions (Continued)},colbacktitle=black,coltitle=white,boxrule=1pt] 
Number of Constraints: <num\_constraints>
\begin{verbatim}
{
    "num_constraints": <num_constraints>,
    "questions": [
        {
            "constraints": [<constraints>], 
            "question": <question>
        },
        ...
    ]
}
\end{verbatim}
~\\
Below is a concrete example:
\\~\\
Number of Constraints: 3
\begin{verbatim}
{
    "num_constraints": 3,
    "questions": [
        {
            "constraints": ["one-way", "New York to London", 
                            "departing next Friday"], 
            "question": "Can you find a one-way flight from New York 
                         to London departing next Friday?"
        },
        ...
    ]
}
\end{verbatim}
\end{tcolorbox}

\begin{tcolorbox}[width=\textwidth,colback={blue!30!},title={Prompt for Iteratively Expanding Constraints and Questions},colbacktitle=black,coltitle=white,boxrule=1pt] 
\textbf{System}:

[Same as above]
\\~\\
\textbf{Instruction}:

[Same as above until ``Your response should follow'']
\\~\\
Your response should follow the JSON format below: 
\\~\\
Maximum number of constraints: <max\_constraints>
\\~\\
Starting constraints and questions:

\begin{verbatim}
{
    "num_constraints": <num_constraints>,
    "constraints": [<constraints>], 
    "question": <question>
}
\end{verbatim}

Questions with increasing complexity:

\begin{verbatim}
{
    "questions": [
        {
            "num_constraints": <starting num_constraints + 1>,
            "constraints": [<previous constraints with 1 additional>], 
            "question": <question>
        },
        {
            "num_constraints": <starting num_constraints + 2>,
            "constraints": [<previous constraints with 2 additional>], 
            "question": <question>
        },
        ... (continue until reaching the maximum number of constraints)
    ]
}
\end{verbatim}

Your Response:
\\~\\
Maximum number of constraints: \{max\_num\_constraints\}
\\~\\
Starting constraints and questions:
\\~\\
\{starting\_constraint\_questions\}
\\~\\
Questions with increasing complexity:
\end{tcolorbox}

\begin{tcolorbox}[width=\textwidth,colback={blue!30!},title={Prompt for Evaluation},colbacktitle=black,coltitle=white,boxrule=1pt] 
\# Interaction Date and Time:
\\~\\
\{interaction\_datetime\}
\\~\\
\# Interaction History:
\newline
\newline
[Concatenation of observations from all steps]
\\~\\
Above are the webpages an assistant interacted with while trying to answer the user's query.
\\~\\
The user is looking for flights with the following constraints:
\\~\\
\{constraints\}
\\~\\
Here is the exact query provided by the user:
\\~\\
\{goal\}
\\~\\
Here is the assistant's response: 
\\~\\
\{message\}
\\~\\
Your task is to evaluate two aspects of the response: 
\\~\\
1) Whether the assistant's response is supported by the interaction history, and 
\\~\\
2) Whether the assistant's response satisfies the user constraints to the extent allowed by the results.
\\~\\
Some Context:
\\~\\
- To determine the seating class of a flight being returned, refer to the value of the "Change seating class" combobox. \\
- It is not always possible to satisfy all the user constraints. In this case, examine whether the response is as close to the user constraints as possible.
\\~\\
Answer in the following format:
\\~\\
<think> \\
Your thoughts and reasoning. \\
</think>
\\~\\
<grounding> \\
Your assessment of whether the response is supported by the interaction history. Answer "yes" or "no" \\
</grounding>
\\~\\
<relevance> \\
Your assessment of whether the response satisfies the user constraints to the extent allowed by the results. Answer "yes" or "no" \\
</relevance>
\end{tcolorbox}

\end{document}